\documentclass{article}

\usepackage[preprint]{nips_2018}

\usepackage{graphicx}
\usepackage{amssymb}
\usepackage{amsmath}
\usepackage[utf8]{inputenc} 
\usepackage[T1]{fontenc}    
\usepackage{hyperref}       
\usepackage{url}            
\usepackage{booktabs}       
\usepackage{amsfonts}       
\usepackage{nicefrac}       
\usepackage{microtype}      
\usepackage{subcaption}
\usepackage{multirow,tabularx}
\usepackage{bbm}
\usepackage{subcaption}
\DeclareMathOperator*{\argmin}{\arg\!\min}
\DeclareMathOperator*{\argmax}{\arg\!\max}
\title{DeepFlow \\ \Large{History Matching in the Space of Deep Generative Models}}

\author{
    Lukas~Mosser \\ Department of Earth Science and Engineering \\
    Imperial College London \\ 
    \texttt{lukas.mosser15@imperial.ac.uk} \\ 
    \And Olivier~Dubrule \\ 
    Department of Earth Science and Engineering \\
    Imperial College London \\ 
    \texttt{o.dubrule@imperial.ac.uk} \\
    \AND Martin J.~Blunt \\ 
    Department of Earth Science and Engineering \\
    Imperial College London \\ 
    \texttt{m.blunt@imperial.ac.uk} \\
}

\begin{document}

\maketitle

\begin{abstract}
  The calibration of a reservoir model with observed transient data of fluid pressures and rates is a key task in obtaining a predictive model of the flow and transport behaviour of the earth's subsurface. The model calibration task, commonly referred to as "history matching", can be formalised as an ill-posed inverse problem where we aim to find the underlying spatial distribution of petrophysical properties that explain the observed dynamic data. We use a generative adversarial network pre-trained on geostatistical object-based models to represent the distribution of rock properties for a synthetic model of a hydrocarbon reservoir. The dynamic behaviour of the reservoir fluids is modelled using a transient two-phase incompressible Darcy formulation. We invert for the underlying reservoir properties by first modeling property distributions using the pre-trained generative model then using the adjoint equations of the forward problem to perform gradient descent on the latent variables that control the output of the generative model. In addition to the dynamic observation data, we include well rock-type constraints by introducing an additional objective function. Our contribution shows that for a synthetic test case, we are able to obtain solutions to the inverse problem by optimising in the latent variable space of a deep generative model, given a set of transient observations of a non-linear forward problem.
\end{abstract}

\section{Introduction}
\label{sec:introduction}
Understanding the flow of fluids within the earth's subsurface is a key issue in many practical applications such as the understanding of how pollutants in aqueous phases are transported and affect the environment \citep{yong1992principles, bear2010modeling}, hydrogeological considerations \citep{domenico1998physical, fetter2018applied}, the formation of mineral deposits \citep{garven1985role, sibson1975seismic}, carbon-capture and sequestration (CCS) \citep{lackner2003guide, holloway2005underground}, and the migration and production of hydrocarbons from subsurface reservoirs \citep{england1987movement}. When modeling subsurface architectures a spatial gridded representation is usually built for modeling physical flow processes \citep{aziz1993reservoir}. The flow of fluids, including aqueous and non-aqueous fluids such as supercritical $CO_2$ is simulated by solving a set of partial differential equations that define the transient behaviour of the flowing liquid phases \citep{gerritsen2005modeling}.
\newpage
In many cases the ability for fluids to flow within a porous material is dominated by two key properties: the effective porosity $\phi$ defines how much pore space is available to flow as a proportion of the bulk volume of the material \citep{bear2013dynamics}. The permeability $\mathbf{k}$ is  a property that is proportional to the flow conductance \citep{darcy1856fontaines, muskat1938flow}. We refer to \cite{blunt2013pore} for a detailed review of the principles of multiphase-fluid flow within porous media. Obtaining direct measurements of these two properties, permeability and porosity, within the earth's structures is difficult and in many cases these measurements are sparse and obtained at various spatial scales, making their integration within the modeling workflow difficult \citep{wu2002analysis, farmer2002upscaling}.

Nevertheless, representing the spatial distribution of the properties of porous media within the earth's subsurface is necessary to build reliable predictive models. In applications such as the injection of supercritical $CO_2$ into hosting rock formations for long-term storage \citep{chadwick2010history, martens2012europe} or the production of hydrocarbons and water from subsurface reservoirs we obtain, in addition to well and seismic data, measurements of the variables dependent on these flow properties ($\mathbf{k}$ and $\phi$) such as fluid phase pressures, produced and injected volumes over time \citep{muskat1981physical}. Calibration of a spatial model of the petrophysical and structural features of the earth's subsurface with available physical measurements is commonly referred to as "history matching" \citep{oliver2011recent}, and more generally as data assimilation \citep{reichle2002hydrologic}. Determining the underlying spatial distributed parameters $\mathbf{x}$ from observed parameters $\mathbf{y}$ such as fluid pressures and rates, can be formalised as an an ill-posed inverse problem.

For an inverse problem to be well-posed it must satisfy three criteria \citep{tikhonov1977solutions}
\begin{enumerate}
    \item Existence: for every set of observed parameters $\mathbf{y}$ there must exist a solution.
    \item Uniqueness: the obtained solution has to be unique.
    \item Stability: the solution is stable with regards to small changes in the observed parameters i.e. a small change in the observed parameters $\mathbf{y}_{obs}$ induces only a small change in the solution parameters $\mathbf{x}$
\end{enumerate}
if these three criteria are not fulfilled the problem is considered ill-posed. It has been shown that inverse problems with spatially distributed parameters common in  groundwater and reservoir flow problems are often ill-posed because they don't fulfill the uniqueness or stability criterion \citep{kravaris1985identification}. \cite{chavent1979identification} provides a detailed analysis of the existence criterion for distributed parameter systems of the heat, transport, and wave equations.


Reducing the number of parameters used to represent the solution space $\mathbf{m}$ allows the inverse problem to be well-defined. \cite{coats1970new} assumed a constant value for large spatial regions of the underlying parameters \citep{cooley1976uniqueness, yehtauxe1971}. \cite{lee1986history} used B-splines to represent spatial distributions of porosity and permeability. Recently, a number of approaches for model parameterisation have been developed based on principal component analysis (PCA) \citep{Vo2014} and improvements thereof using kernel and optimisation-based PCA. \cite{demyanov2011reservoir} use kernel learning to find a parameterisation of the petrophysical parameter space.

In a Bayesian setting of the inverse problem \citep{tarantola2005inverse} the gradual deformation method applies a continuous perturbation of a set of model parameters $\mathbf{m}$ obtained from a prior distribution to honor the observed data of the forward problem \citep{roggero1998gradual,Caers2007}. Bayesian approaches to solve the reservoir history matching problems, such as the Ensemble Kalman Filter (EnKF) \citep{evensen1994sequential,evensen2003ensemble} have been applied in numerous history-matching case studies \citep{lorentzen2005analysis}. \cite{emerick2012history} present an ensemble method based on EnKF, that shows improved performance in the case of highly non-linear problems.

Recent developments in deep learning \citep{lecun2015deep, goodfellow2016deep} have motivated the use of neural-network-based parameter representations. \cite{arauco18} parameterised geological training images using a Deep Belief Network (DBN) \citep{hinton2006fast, hinton2006reducing} and used an ensemble smoother with multiple data assimilation (ES-MDA) \citep{emerick2013ensemble} to perform history matching. Variational autoencoders \citep{LALOY2017387} and spatial generative adversarial networks (SGAN) \citep{jetchev2016,laloy2018training} have been used to generate a set of channelised training images to perform probabilistic inversion using adaptive Markov-Chain Monte-Carlo methods \citep{laloy2012high}. \cite{chan2018exemplar} used generative adversarial networks trained using patch-based kernel discrepancy measures to synthesise geological models from training images. Realisations conditional to spatial observations of the underlying model parameters can be obtained by gradient-based image inpainting methods \citep{yeh2016inpainting, mosser2018, dupont2018generating} or direct learning of conditional image distributions using autoregressive generative models such as PixelCNNs \citep{van2016conditional,2018arXiv181003728D}. 

Adjoint-state methods allow gradients of the forward problem with respect to the solution parameters $\mathbf{m}$ to be obtained. These gradients can then be further backpropagated through a deep generative models such as generative adversarial networks \citep{goodfellow2014, kingma2013auto, modelorder, mosser2018stochastic} to perform inversion in the space of latent variable models and have been used for probabilistic inversion in the case of acoustic waveform inversion using a Metropolis-adjusted Langevin algorithm (MALA) \citep{roberts2002langevin, mosser2018stochastic}. Numerous studies have evaluated the use of deep generative models such as GANs for linear inverse problems \citep{shah2018solving} in the context of compressed sensing \citep{bora2018ambientgan, mardani2019deep} and have analysed the convergence of inversion algorithms that  use GANs \citep{bora2017compressed, shah2018solving}.

A number of recent approaches have combined methods developed in the context of deep neural networks with traditional approaches to solve (ill-posed) inverse problems such as learning the regularisation itself \citep{li2018nett,lunz2018adversarial}, using iterative deep neural networks \citep{adler2017solving}, or introducing deep neural networks that account for non-linear forward operators in reconstruction algorithms \citep{adler2018learned}.

Following the work of \citet{mosser2018stochastic} we parameterise a family of geological models and associated permeability and porosity values using a GAN. We solve the transient two-phase immiscible Darcy flow equations without gravity and under isothermal conditions using a finite-difference approach and obtain gradients of the mismatch between observed and simulated data with respect to the gridblock permeability and porosity using the adjoint-state method. 

Starting from random initial locations within the underlying Gaussian latent space controlling the GAN's output, we use gradient-descent to minimise the misfit between the flow data and the forward model output by modifying the set of latent variables. A regularisation scheme of the optimisation problem is derived from a Bayesian inversion setting. Based on a two-dimensional synthetic case of a channelised reservoir system we show that gradient-based inversion using GANs as a model parameterisation can be achieved, honoring the observed physical quantities of the non-linear forward problem as well as borehole measurements of the underlying petrophysical quantities. The trained models and code are available as open-source software\footnote{\hyperlink{https://github.com/LukasMosser/DeepFlow}{https://github.com/LukasMosser/DeepFlow}}.

\section{Methodology}
\label{sec:methodology}
We consider the problem of determining the flow behaviour of a subsurface hydrocarbon reservoir from which oil and water are being produced from a well. To displace the oil phase we inject water at a constant injection rate in another well and produce oil by maintaining a constant bottom-hole pressure. The forward problem $\mathcal{F}(\cdot)$ is therefore defined as solving the two-phase (oil and water) Darcy flow equations with a slightly compressible oil phase with no gravity effects and under isothermal conditions. The observed variables of the system $y_{obs}$ are the injection pressure $p_{inj}(\mathbf{x}_i, t)$ at the location of the injection well $\mathbf{x}_i$ as a function of time $t$, as well as the produced volumes of water $q_{w}(\mathbf{x}_p, t)$ and oil $q_{o}(\mathbf{x}_p, t)$ at the production well $\mathbf{x}_p$.
\begin{equation}
\begin{gathered}
    \mathbf{y}_{obs, flow} = \{p_{inj}(\mathbf{x}_i, t), q_{w}(\mathbf{x}_p, t), q_{o}(\mathbf{x}_p, t)\} \\
    \mathbf{y}_{obs, well} = \{\mathbbm{1}_r(\mathbf{x}_{wells})\}
    \label{equ:observed_variables}
\end{gathered}
\end{equation}
We consider all fluid parameters of the forward problem, the oil and water viscosity $\mu_{o, w}$, density $\rho_{o, w}$, compressibility $c_{o, w}$, as well as the relative-permeability behaviour, represented by a Brooks-Corey model, as constants \citep{blunt2017multiphase}. A table of fluid properties is provided in the appendix (see table~\ref{tab:simulation_parameters}). Therefore, the set of model parameters $\mathbf{m}(\mathbf{x})$ are the spatial distribution of the reservoir rock type indicator function $\mathbbm{1}_r(\mathbf{x})$, the single-phase permeability $\mathbf{k}(\mathbf{x})$, and porosity $\phi(\mathbf{x})$.
\begin{equation}
    \mathbf{m}(\mathbf{x}) = \{\mathbbm{1}_r(\mathbf{x}), \mathbf{k}(\mathbf{x}), \phi(\mathbf{x})\}
    \label{equ:model_parameters}
\end{equation}
We assume a binary set of reservoir rock-types where $\mathbbm{1}_r(\mathbf{x})=0$ corresponds to a low permeability shale and $\mathbbm{1}_r(\mathbf{x})=1$ indicates the high permeability sandstone of a river-channel.

Given the non-linear forward operator $\mathcal{F}(\cdot)$ this allows us to define the forward problem as
\begin{subequations}
\begin{equation}
     \mathbf{y}_{\mathbf{m}} = \mathcal{F}(\mathbf{m})
\end{equation}
\begin{equation}
     \mathbf{y}_{obs, flow} = \mathbf{y}_{\mathbf{m}} + \mathbf{\varepsilon}, \  \mathbf{\varepsilon}\sim\mathcal{N}(\mathbf{\mu}_{\epsilon}, \mathbf{\sigma}^2_{\epsilon}\mathbf{I})
\end{equation}
\label{equ:forward_problem}
\end{subequations}

We represent the distribution of possible model parameters $\mathbf{m}$ by a (deep) generative latent-variable model
\begin{equation}
\begin{gathered}
    \mathbf{m} = G_{\mathbf{\theta}}(\mathbf{z}) \\
    \mathbf{z} \sim p(\mathbf{z})=\mathcal{N}(0, \mathbf{I})
    \label{equ:generative}
\end{gathered}
\end{equation}
with latent variables $\mathbf{z}$ sampled from a Gaussian distribution (Eq.~\ref{equ:generative}).
Here we assume that the generative model has been trained prior to the inversion process, and is therefore constant.

Furthermore we can write:
\begin{equation}
    \mathbf{y}_{obs, flow} = \mathbf{y}_{\mathbf{m}} + \mathbf{\varepsilon} = \mathcal{F}(G_{\mathbf{\theta}}(\mathbf{z}))+ \mathbf{\varepsilon}
    \label{equ:forward_problem_gan}
\end{equation}
and from Bayes theorem
\begin{equation}
    p(\mathbf{z}|\mathbf{y}_{obs, flow}) \propto p(\mathbf{y}_{obs, flow} | \mathbf{z})p(\mathbf{z})
    \label{equ:propto}
\end{equation}

 where for the observed flow data $\mathbf{y}_{obs, flow}$ the likelihood function $p(\mathbf{y}_{obs, flow} | \mathbf{z})$ is a Gaussian with mean $\mathcal{F}(G_{\mathbf{\theta}}(\mathbf{z}))$ and variance $\mathbf{\varepsilon}$. We have assumed  a Bernoulli likelihood for the observed facies at the wells (Eq.~\ref{equ:model_parameters}). 
 
 Furthermore, we assume conditional independence of the likelihood of the observed flow and facies data at the wells 
 \begin{equation}
    p(\mathbf{y}_{obs} | \mathbf{z}) = p(\mathbf{y}_{obs, flow} | \mathbf{z})p(\mathbf{y}_{obs, well} | \mathbf{z})
    \label{equ:independence}
\end{equation}
 
 In the case of GANs only individual samples can be obtained and therefore estimation of the posterior is restricted to point-estimates. We derive a regularisation scheme for the history matching problem from finding point estimates of the maxima of the posterior distribution (MAP)
\begin{subequations}\label{equ:map}
\begin{gather}
\argmax_{\mathbf{z}}\{p(\mathbf{y}_{obs, flow}|\mathbf{z})p(\mathbf{y}_{obs, well}|\mathbf{z})p(\mathbf{z})\} \\
=\argmax_{\mathbf{z}}\{\log p(\mathbf{y}_{obs, flow}|\mathbf{z})+\log p(\mathbf{y}_{obs, well}|\mathbf{z})+\log p(\mathbf{z})\} \\
=\argmin_{\mathbf{z}}\{-\log p(\mathbf{y}_{obs, flow}|\mathbf{z})-\log p(\mathbf{y}_{obs, well}|\mathbf{z})-\log p(\mathbf{z})\} \\
=\argmin_{\mathbf{z}}\{\mathcal{L}_{flow}(\mathbf{z})-\mathcal{L}_{wells}(\mathbf{z})-\mathcal{L}_{prior}(\mathbf{z})\} \\
=\argmin_{\mathbf{z}}\{\frac{1}{2\sigma_{\epsilon}^2}\|\mathbf{y}_{obs, flow}-\mathcal{F}(G_{\mathbf{\theta}}(\mathbf{z}))\|^2_2 \\
-\sum^{\substack{Well\\Cells\\}}\left[\mathbbm{1}_r(\mathbf{x}_{wells})\log(p(\mathbbm{1}_r(\mathbf{x}_{wells})))+(1-\mathbbm{1}_r(\mathbf{x}_{wells}))\log(1-p(\mathbbm{1}_r(\mathbf{x}_{wells})))\right] \nonumber \\
-\log p(\mathbf{z})\}\nonumber
\end{gather}
\end{subequations}
where the squared-error norm of the difference between the observed data $\mathbf{y}_{obs, flow}$ and the modelled data $\mathbf{y}_{\mathbf{m}}=\mathcal{F}(G_{\mathbf{\theta}}(\mathbf{z}))$ is a result of the assumption of a Gaussian likelihood in equation~\ref{equ:forward_problem} and the well loss $\mathcal{L}_{wells}(\mathbf{z})$ corresponds to the binary cross-entropy between the observed facies indicator $\mathbbm{1}_r(\mathbf{x}_{wells})$ and the facies probability $p(\mathbbm{1}_r)$ of each grid-block at the wells.

The derived regularisation scheme can be framed as a minimisation problem over the latent variables i.e. we minimise Eq.~\ref{equ:map} by modifying the latent variables $\mathbf{z}$. Specifically we use a squared-error functional that measures the difference between the observed pressures and rates at the well locations averaged over the total duration $T$ of the observed data 
\begin{equation}
\begin{split}
\mathcal{L}_{flow}(\mathbf{z})=\frac{1}{T}\sum_{t=0}^{T}\Biggl[\left(\frac{\|q_{w}^{obs}(\mathbf{x}_p, \ t)-q_w^{\mathbf{m}}(\mathbf{x}_p, \ t)\|}{\sigma_q}\right)^2 \\ 
+ \left(\frac{\|q_o^{obs}(\mathbf{x}_p, \ t)-q_o^{\mathbf{m}}(\mathbf{x}_p, \ t)\|}{\sigma_q}\right)^2 \\ 
+ \left(\frac{\|p^{obs}(\mathbf{x}_i, \ t)-p^{\mathbf{m}}(\mathbf{x}_i, \ t)\|}{\sigma_p}\right)^2\Biggr]
\label{equ:functional}
\end{split}
\end{equation}
where $\mathbf{x}_{p}$ and $\mathbf{x}_{i}$ are the locations of the production and injection wells respectively and measurement uncertainties have been quantified by the standard deviation of the oil and water rate
\begin{equation}
\sigma_q = 0.03~q_{inj}
\label{equ:q_t}
\end{equation}
and the standard deviation of the pressure measurements
\begin{equation}
    \sigma_p = 0.05~p_{ref}
    \label{equ:max_p}
\end{equation}
used to generate the set of observed pressure and rate data.

Using the adjoint-state method \citep{plessix2006review, suwartadi2012nonlinear, krogstad2015mrst} and traditional neural network backpropagation \citep{rumelhart1988learning} we can obtain gradients $\frac{\partial{\mathcal{L}_{flow}(\mathbf{z})}}{\partial{\mathbf{z}}}$ of the error functional $\mathcal{L}_{flow}(\mathbf{z})$ (Eq.~\ref{equ:functional}) with respect to the latent variables.

\begin{figure}[!htb]
    \centering
    \includegraphics[width=\textwidth]{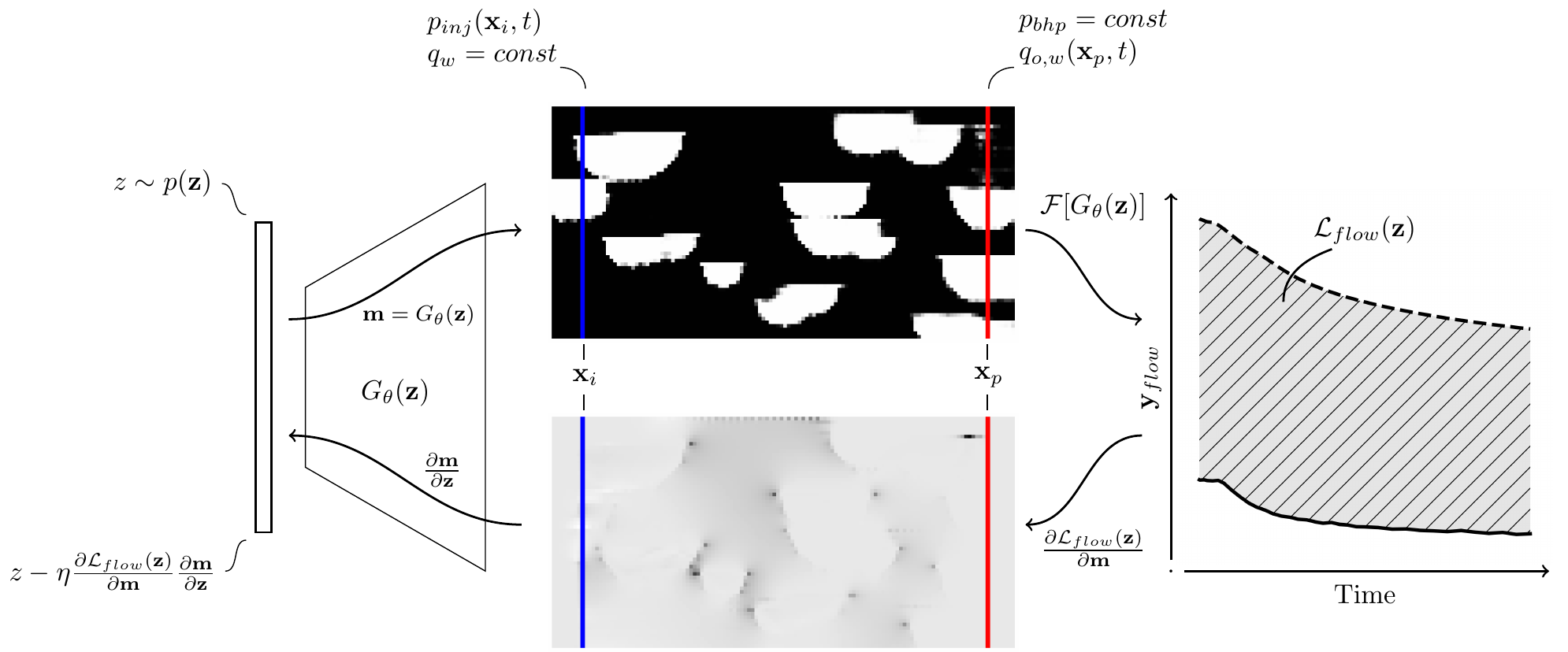}
    \caption{Overview of a single forward and backward-pass through the combined generative network $G_{\mathbf{\theta}}$ and forward problem of two-phase Darcy flow $\mathcal{F}(\cdot)$. A set of model parameters $\mathbf{m}=G_{\mathbf{\theta}}(\mathbf{z})$ is generated from a forward-pass through the generative model $G_{\mathbf{\theta}}$. The generative model creates continuous spatial properties such as the permeability $\mathbf{k}$ shown here (top). The forward problem is solved where water is injected in the injection well (blue) and oil and water are produced from the production well (red). The error functional between observed parameters $\mathbf{y}_{obs}$ and modelled observations $\mathbf{y}_{\mathbf{m}}=\mathcal{F}(\mathbf{m})=\mathcal{F}(G_{\mathbf{\theta}}(\mathbf{z}))$ is evaluated. A spatial distributed gradient (bottom) with respect to the underlying properties is computed using the adjoint state method. The obtained gradients are then backpropagated by traditional neural-network backpropagation to compute a gradient with respect to the latent variables. The error functional is minimised by performing gradient descent using the obtained gradient.}
    \label{fig:overview_workflow}
\end{figure}

We therefore seek to minimise the total loss
\begin{equation}
    \mathcal{L}(\mathbf{z}) = \mathcal{L}_{flow}(\mathbf{z}) + \mathcal{L}_{well}(\mathbf{z}) + \mathcal{L}_{prior}(\mathbf{z})
    \label{equ:total_loss_flow_prior}
\end{equation}
consisting of the flow loss $\mathcal{L}_{flow}(\mathbf{z})$ (Eq.~\ref{equ:functional}), the well-data loss $\mathcal{L}_{wells}(\mathbf{z})$, and the prior loss $\mathcal{L}_{prior}(\mathbf{z})=-\log{p(\mathbf{z})}$ (Eq.~\ref{equ:map}).
We note that this approach is similar to the approach by \cite{creswell2018inverting} which includes an additional loss term corresponding to the prior distribution of the latent variables when inverting the generator function of a GAN. The system of partial differential equations of the non-linear forward problem as well as the likelihood function (Eq.~\ref{equ:functional}) have been previously implemented in the open-source MATLAB/GNU-Octave reservoir simulation framework MRST \citep{lie_2019}. An overview of the end-to-end coupling between the latent variables, the deep generative model $G_{\mathbf{\theta}}$, and the numerical forward model $\mathcal{F}(\cdot)$ is shown in figure~\ref{fig:overview_workflow}.

\begin{figure}[!htb]
    \centering
    \includegraphics[width=\textwidth]{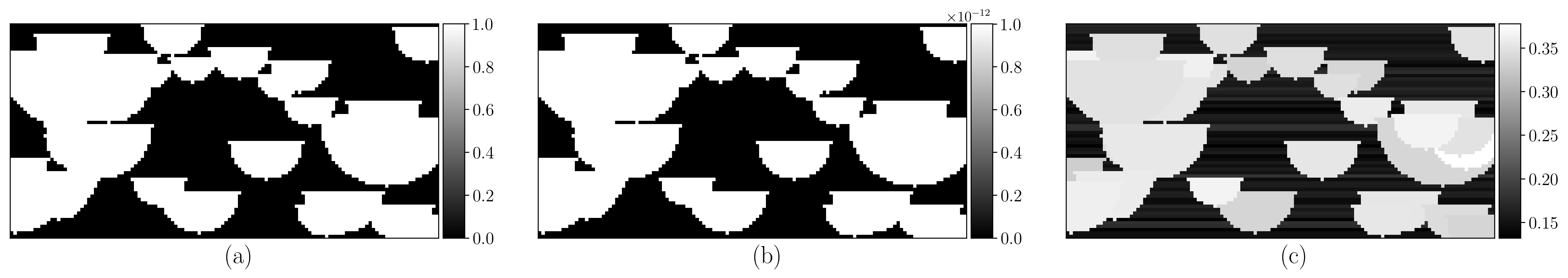}
    \caption{Object-based model used to generate reference observed data. Each object-based model has three associated properties: the rock-type (shale 0 - sandstone 1) indicator $\mathbbm{1}_r$ (a), the grid block permeability $\mathbf{k}$ (b), and porosity $\phi$ (c). The model was obtained from a test-set of object-based model realisations that the generator network $G_{\mathbf{\theta}}(\mathbf{z})$ was not trained on.}
    \label{fig:gt_model}
\end{figure}

To evaluate the proposed approach we create synthetic two-dimensional vertical cross-sections of a stacked river-channel system using an object-based approach. Channel-bodies are represented by half-circles with centers following uniform distributions in space. The full object-based modeling workflow is detailed in \cite{mosser2018stochastic}. For the case of reservoir history matching (Eq.~\ref{equ:forward_problem}) each object-based model is associated with the three model parameters $\mathbf{m}$ following equation~\ref{equ:model_parameters} (Fig.~\ref{fig:gt_model}). 
\newpage
We train a deep generative adversarial network (GAN) \citep{goodfellow2014} on a set of $1\times10^4$ object-based realisations to learn the generative model $G_{\mathbf{\theta}}(\mathbf{z})$ (Eq.~\ref{equ:generative}) prior to obtaining MAP samples (Eq.~\ref{equ:map}) by optimising equation~\ref{equ:total_loss_flow_prior}. Specifically, we train a so-called Wasserstein-GAN \citep{arjovsky2017wasserstein,gulrajani2017improved}, that minimises the Wasserstein distance between generated and real probability distributions. To stabilise training of the Wasserstein-GAN we use the Lipschitz penalty introduced by \cite{2017arXiv170908894P}. A detailed overview of the training procedure for the deep-generative model is presented in \cite{mosser2018stochastic}. Network architectures for the generator $G_{\mathbf{\theta}}(\mathbf{z})$ and discriminator $D_{\mathbf{\omega}}$ are detailed in appendix table~\ref{tab:gan_architecture}.

The generator network $G_{\mathbf{\theta}}(\mathbf{z})$ is a deep convolutional neural network that maps a 100-dimensional noise-sample from a standardised multi-Gaussian distribution $z\sim\mathcal{N}(0, \mathbf{I})$ to a spatial distribution of the model parameters $\mathbf{m}$ (Eq.~\ref{equ:model_parameters}) with 128 pixels in the x- and 64 pixels in the z-direction
\begin{equation}
\mathbb{R}^{50\times2\times1}\rightarrow\mathbb{R}^{3\times128\times64}
\label{equ:mapping}
\end{equation}
where the three output features $\mathbb{R}^{3\times128\times64}$ correspond to the grid-block rock-type probability, permeability and porosity (Eq.~\ref{equ:model_parameters}). Due to the continuous nature of the generative networks sigmoid activation function, the output of the generative network corresponding to a representation of the reservoir rock-type $\mathbbm{1}_r$ can be interpreted as a rock-type probability. The grid-block permeability and porosity values are obtained by transforming the generated rock-type probability from the interval $(0-1)$ to a permeability by a linear transformation (see appendix table~\ref{tab:generator}) to rescale the output of the generative model to a range of values that correspond to admissible petrophysical values for sandstone and shale. The finite-difference approximation of the forward model is computed on the regular grid given by the pixel-based output of the generative model $G_{\mathbf{\theta}}(\mathbf{z})$ and hence each pixel-based property corresponds to the grid-block property used in the numerical evaluation of the forward problem. 

We have selected a reference realisation from a test-set of object-based realisations that were generated independently of the models used to train the generative network $G_{\mathbf{\theta}}$. The model shown in figure~\ref{fig:gt_model} was used to generate a set of synthetic observed data $\mathbf{y}_{obs}$ that served as a reference case used to validate the proposed inversion approach. Gaussian noise was added to the production rates and pressures based on assumed measurement uncertainties (Eq.~\ref{equ:q_t}-\ref{equ:max_p}). The transient pressure and rates of oil and water were recorded for a duration of 600 days and the full history was used in the gradient-based optimisation outlined in equation~\ref{equ:map}. To minimise the total loss (Eq.~\ref{equ:total_loss_flow_prior}) we used the ADAM optimiser \citep{kingma2014adam} using a fixed step size and $\beta$-parameters ($\eta=3\times10^{-2}$, $\beta_1=0.9$, $\beta_2=0.999$). The simulation parameters used for the numerical evaluation of the forward problem are provided in appendix table~\ref{tab:simulation_parameters}.
\newpage
\section{Results}
\label{sec:results}
\begin{figure}
    \centering
    \includegraphics[width=\textwidth]{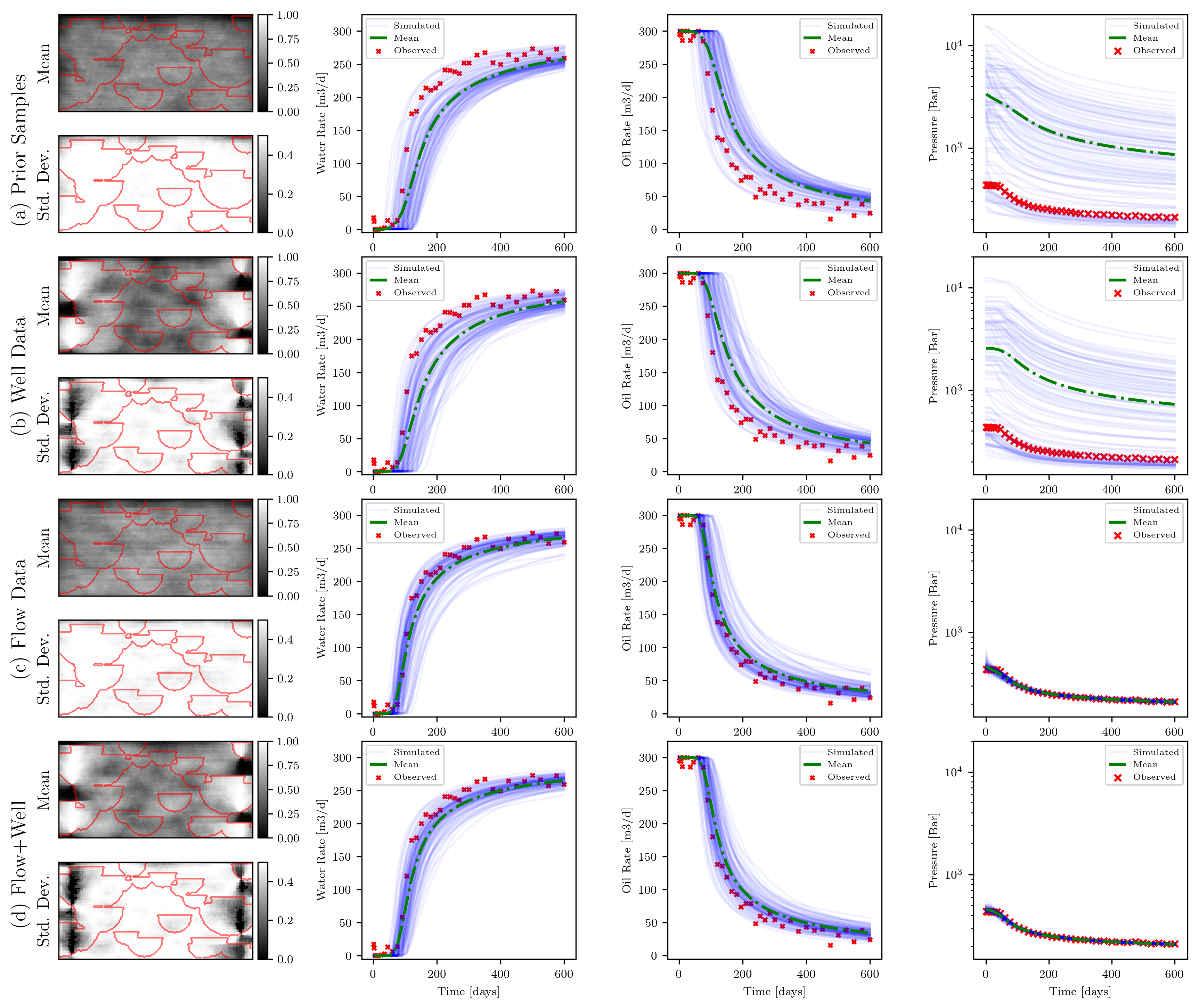}
    \caption{(a) Overview of the dynamic flow behaviour, as well as average and standard deviation of the gridblock rock-types for $N=100$ samples obtained from the unconditional prior distribution i.e. samples drawn at random from the generative model. (b) Model realisations after optimising only the gridblock rock-type at wells, (c) optimising only the dynamic flow behaviour, and (d) optimising both well rock-type and flow behaviour. Red contours on the mean and standard deviation of rock-type distributions reflect the reference model of object-based channel-bodies shown in figure~\ref{fig:gt_model}.}
    \label{fig:runs_comparison}
\end{figure}
Four history matching scenarios were considered:
\begin{itemize}
    \item Scenario 1: Samples of the unconditional prior were obtained ($N=1000$) and a single evaluation of the forward problem was performed. This allowed us to evaluate the prior distribution of the dynamic flow behaviour of the underlying generative model $\mathbf{y}_{prior }=\mathcal{F}(\mathbf{m})=\mathcal{F}[G_{\mathbf{\theta}}(\mathbf{z})]$ (Fig.~\ref{fig:runs_comparison}-a).
    \item Scenario 2: Inversion was performed by considering only the well data loss $\mathcal{L}_{well}(\mathbf{z})$ at the injection and production wells and the prior loss $\mathcal{L}_{prior}(\mathbf{z})$ to honor the prior distribution of the latent variables $\mathbf{z}$  (Fig.~\ref{fig:runs_comparison}-b).
    \item Scenario 3: We performed inversion by considering only the flow-based loss $\mathcal{L}_{flow}(\mathbf{z})$ due to the mismatch between simulated and observed dynamic two-phase flow data, as well as the prior loss on the latent variables $\mathcal{L}_{prior}(\mathbf{z})$ (Fig.~\ref{fig:runs_comparison}-c).
    \item Scenario 4: All three losses were combined: the flow loss $\mathcal{L}_{flow}(\mathbf{z})$, the well-data loss $\mathcal{L}_{well}(\mathbf{z})$, and the prior loss $\mathcal{L}_{prior}(\mathbf{z})$, corresponding to the total loss $\mathcal{L}(\mathbf{z})$, as outlined in equation~\ref{equ:total_loss_flow_prior} (Fig.~\ref{fig:runs_comparison}-d).
\end{itemize} 
For scenario 2 to 4, we performed $N=100$ inversions, starting from the same set of $100$ initial random samples of latent variables $\mathbf{z}$. Each optimisation procedure was run for a total of 500 ADAM optimisation steps. For scenarios 3 and 4, we choose the iteration with the lowest total loss as the output of each run. Early stopping was performed in scenario 2 where only the well data loss was considered when the facies accuracy at the well reached 100$\%$.  
\begin{figure}
    \centering
    \includegraphics[width=\textwidth]{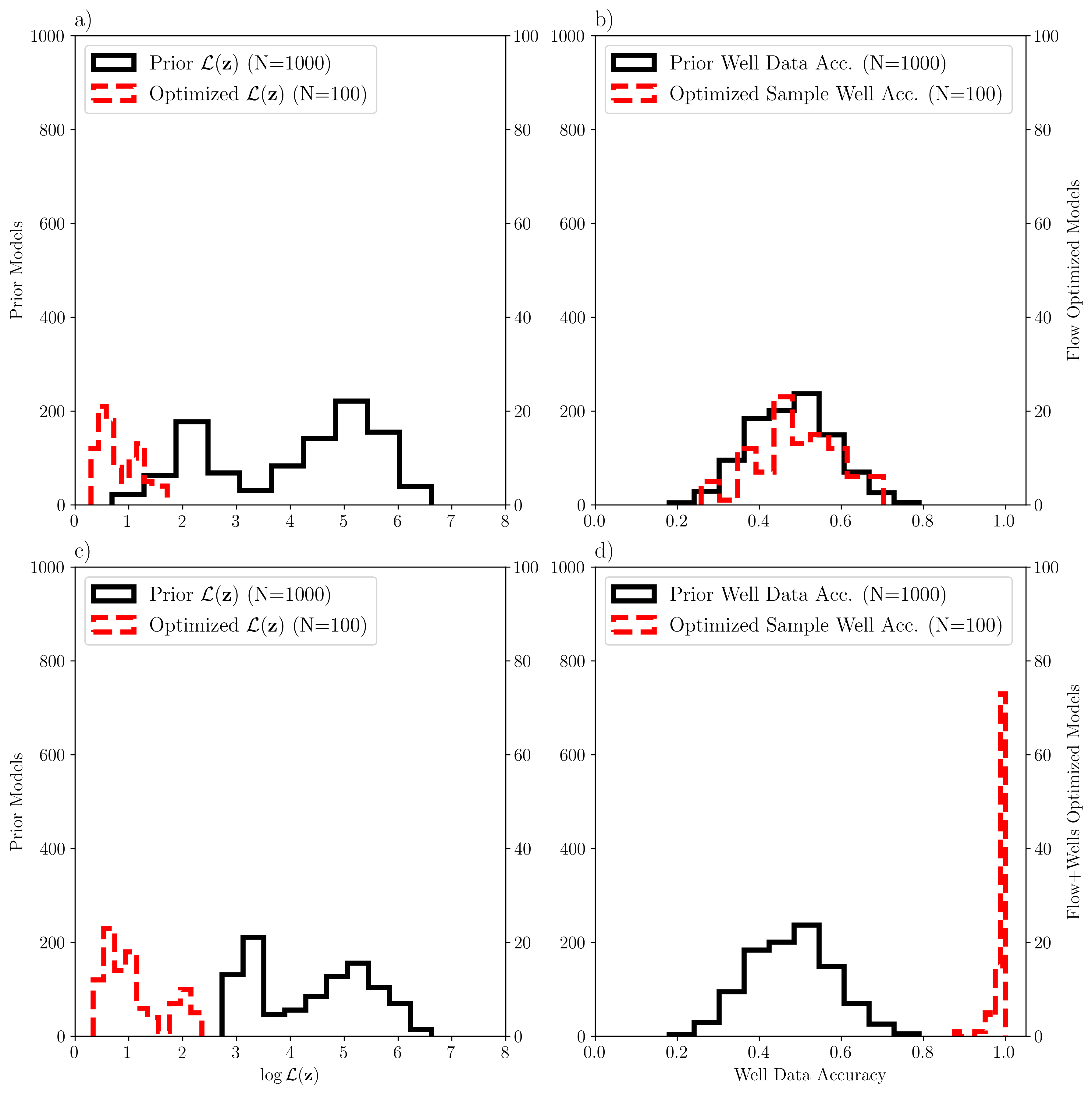}
    \caption{Histograms of the total loss functional $\log{\mathcal{L}(\mathbf{z})}$ (a, c) and the grid-block rock-type accuracy at the wells (b, d). Models ($N=100$) optimised on the observed flow data only (a-b, red) and optimised on both flow and well rock type (c-d, red) are compared to models ($N=1000$) sampled unconditionally from the prior distribution (a-d, black). In both cases, we observe that optimising the flow-based objective leads to realisations that match the observed data closer than sampling unconditionally from the prior. The accuracy of the well rock type  for models optimised only on the dynamic flow data follow the prior distribution of models, whereas the models obtained by optimising flow and well-data loss functions have high accuracy at the well locations.}
    \label{fig:histogram_comparison}
\end{figure}
Figure~\ref{fig:runs_comparison} shows a comparison of these four scenarios in terms of their dynamic flow behaviour i.e. pressures and rates at production and injection wells as a function of the simulation time, as well as the mean and standard deviation of the thresholded rock-type probability maps. The threshold for rock-type probability maps was set at 0.5 to distinguish between sand (1) and shale (0).

The prior distribution of the observed variables $\mathbf{y}_{prior}$ (Scenario 1, Fig.~\ref{fig:runs_comparison}-a, right) shows a large spread in the distribution of the observed oil-water production rates and injection pressure. The observed liquid rates of the reference case (red) lie at the edge of the unconditional data distribution. The pressure behaviour of the prior models varies across 2 orders of magnitude in relation with the pressure behaviour of the reference case at the low end of the pressure versus time distribution.  

The unconditional prior samples of the model parameters, shown here as the rock-type indicator function $\mathbbm{1}_r$, are shown as cross-sections of the mean and standard deviation of the $N=100$ models (Scenario 1, Fig.~\ref{fig:runs_comparison}-a, left). As shown by the standard deviation map, nearly all values are close to a value of 0.5 as expected from an indicator function of mean 0.5. 

Considering the likelihood of the grid-block rock-type indicator function at the wells only (Scenario 2) where the binary cross-entropy loss at the injection and production well $\mathcal{L}_{well}(\mathbf{z})$, as well as the prior loss $\mathcal{L}_{prior}(\mathbf{z})$ were minimised, we find a large spread in the production behaviour of the generated samples closely resembling the unconditional prior distribution (Scenario 2, Fig.~\ref{fig:runs_comparison}-b, right). Inspecting the mean and standard deviation maps of the rock-type indicator function of the inverted samples matching the well data only, we find that these samples match the well data while their flow behaviour is consistent with the unconditional prior. This means that constraining only by well data has very little effect on the facies model connectivity between the wells. A lateral effect of the grid-block well data can be observed comparing the outline of the mean rock-type maps and the red contours of the ground truth model (Fig.~\ref{fig:gt_model}) used to generate the reference observed data $\mathbf{y}_{obs, flow}$ and $\mathbf{y}_{obs, well}$.

When only the flow loss $\mathcal{L}_{flow}(\mathbf{z})$ was considered in the optimisation, the production data of the generated realisations were tightly constrained to the observed liquid rates and pressures (Scenario 3, Fig.~\ref{fig:runs_comparison}-c, right). The mean and standard deviation maps for the obtained distribution of the rock-type indicator functions $\mathbbm{1}_r$ do not indicate any clear structural features and resemble the prior maps shown in the first row of figure~\ref{fig:runs_comparison}. We have obtained samples of model parameters $\mathbf{m}$ that all honor the observed dynamic flow data $\mathbf{y}_{obs, flow}$, but which are not anchored at the wells by the observed well rock types.
\begin{figure}
    \centering
    \includegraphics[width=\textwidth]{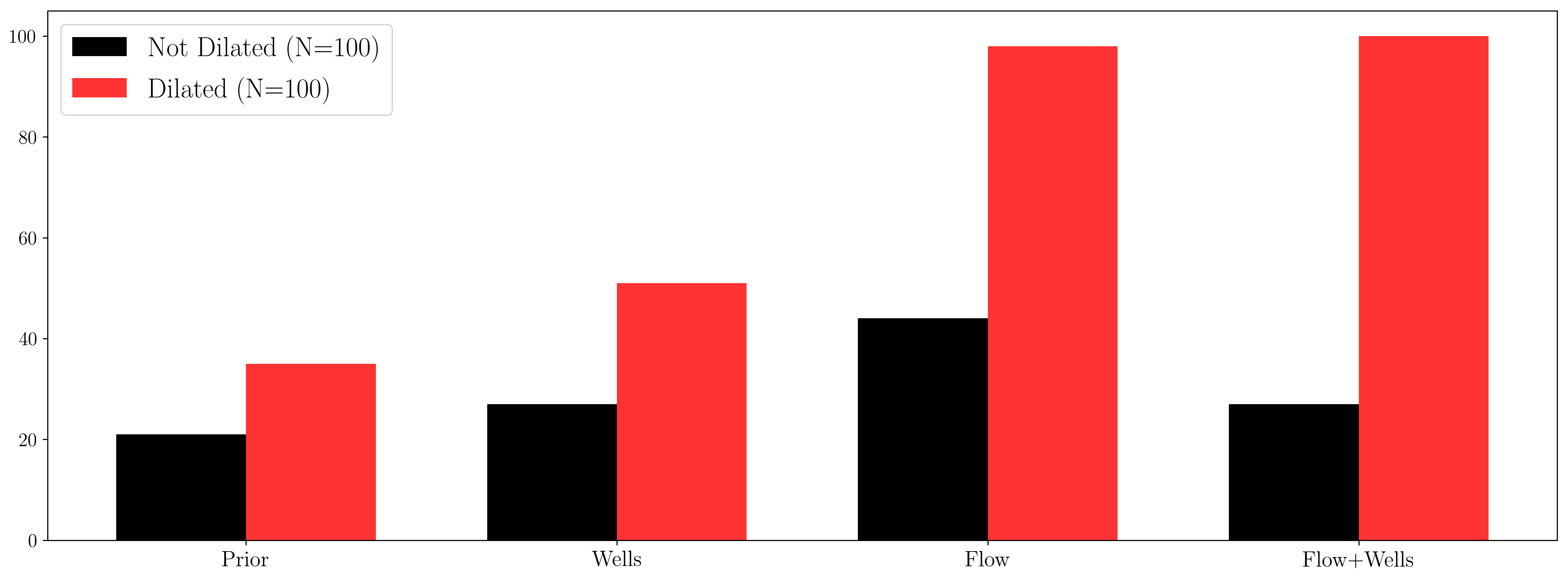}
    \caption{Comparison of the number of models that have a connected cluster of channel-bodies that connects the injection and production wells in each individual sample obtained in the inversion process for the four scenarios outlined in section~\ref{sec:methodology}. After Boolean dilation, nearly all models that were optimised with respect to the observed dynamic flow data form a connected cluster between the two wells, whereas half of the samples obtained from honoring only the well data remained disconnected.}
    \label{fig:connected_components}
\end{figure}
Combining the flow loss $\mathcal{L}_{flow}(\mathbf{z})$, the well data loss $\mathcal{L}_{well}(\mathbf{z})$ and prior loss $\mathcal{L}_{prior}(\mathbf{z})$ as the total loss (Eq.~\ref{equ:total_loss_flow_prior}) (Scenario 4, Fig.~\ref{fig:runs_comparison}-d) we obtain samples that match the observed dynamic behaviour closer than scenario three, which only considered the flow loss. In addition to honoring the dynamic observed data $\mathbf{y}_{obs, flow}$, the facies data $\mathbbm{1}_r$ of the obtained samples at the wells match the observed grid-block data at the production and injection well $\mathbf{y}_{obs, well}$. This is indicated by the low standard deviation map of the $N=100$ inverted models (Scenario 4, Fig.~\ref{fig:runs_comparison}-d, left).
\begin{figure}[!htb]
    \centering
    \includegraphics[width=\textwidth]{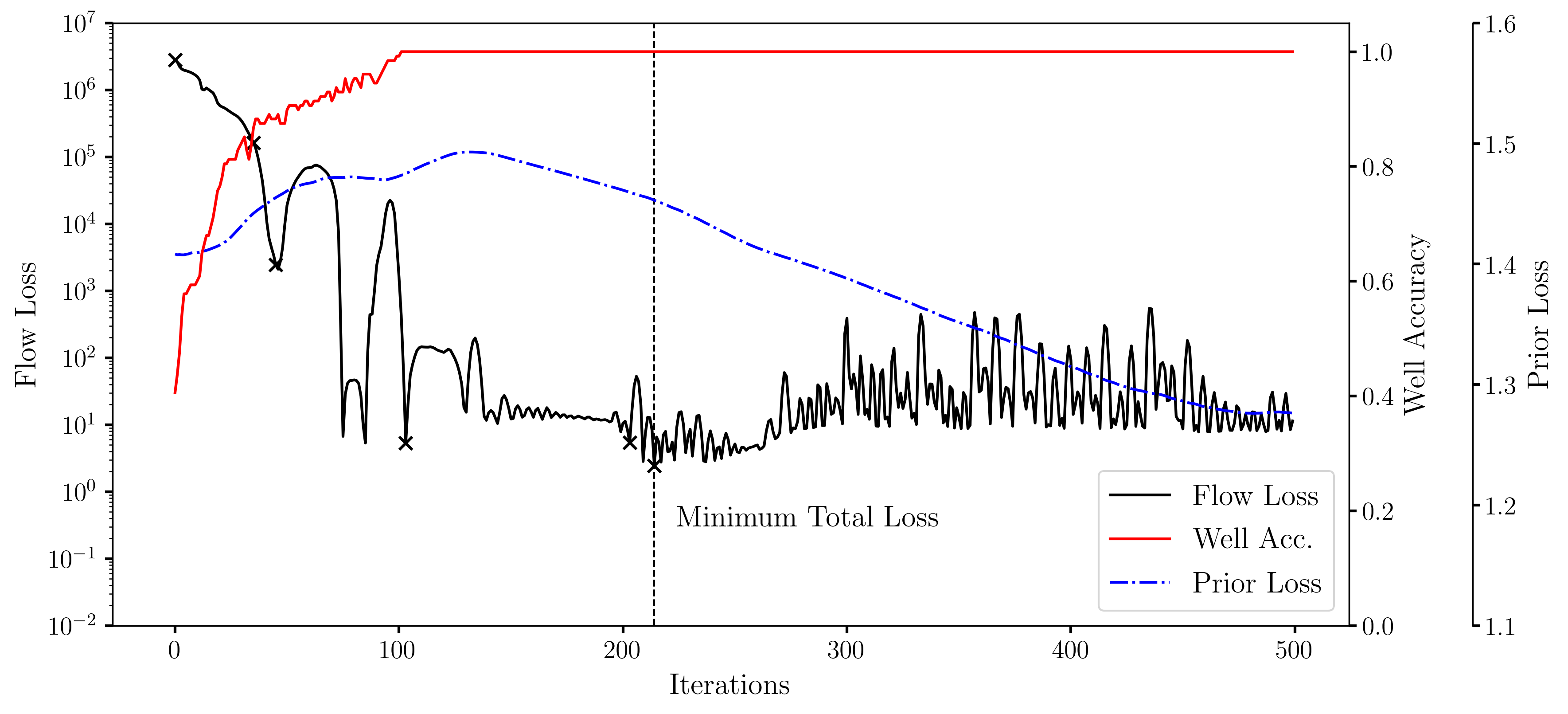}
    \caption{Optimisation trajectory for a selected inverted sample based on flow, well-data and prior losses. The flow loss $\mathcal{L}(\mathbf{z})$ shows a highly non-linear optimisation problem with many local minima leading up to the smallest error being achieved after 214 iterations. The well data accuracy at the production and injection wells quickly reaches values greater than 90\% accuracy. Cross-marks indicate the loss values of the intermediate realisations presented in figure~\ref{fig:facies_flowwells_case_66}.}
    \label{fig:functional_flowwells_case_66}
\end{figure}
Figure~\ref{fig:histogram_comparison} compares the distribution of the total loss ($\mathcal{L}(\mathbf{z})$, Eq.~\ref{equ:total_loss_flow_prior}) for the $N=1000$ models obtained from the unconditional prior distribution i.e. sampling a set of model parameters $\mathbf{m}$ (Eq.~\ref{equ:model_parameters}) from the generative model $G_{\mathbf{\theta}}(\mathbf{z})$ and solving the forward problem once, (Eq.~\ref{equ:forward_problem}) and the models of scenario 3 (flow and prior loss only) and 4 (flow, well and prior loss). For both considered scenarios (Fig.~\ref{fig:histogram_comparison}, a and c respectively) the distribution of the total loss is shifted to lower values when compared to the total loss distribution of the models obtained by sampling from the prior distribution. 
The distribution of well rock type accuracy shows that history matching only flow data (Fig.~\ref{fig:histogram_comparison} b) does not improve the rock type accuracy. When the well-data loss $\mathcal{L}_{well}(\mathbf{z})$ is optimised in addition to the flow-loss (Fig.~\ref{fig:histogram_comparison} d), we observe that the obtained models closely match the rock type at the wells in addition to the dynamic flow data.

Due to the high computational cost of the forward model, reaching a good match with the observed data, within a reasonable number of forward-model evaluations, is necessary. Figure~\ref{fig:error_threshold_flowwells} shows the distribution of the number of optimisation iterations of equation~\ref{equ:map} necessary to reach a given threshold value of the total loss $\mathcal{L}(\mathbf{z})$ for the case where all three objective functions are used in the history matching process (Scenario 4).  All $N=100$ inverted models achieve a total-loss less than $1\times10^{3}$ (Fig.~\ref{fig:error_threshold_flowwells} c) which corresponds to the low end of the loss values obtained by sampling from the prior distribution only (Fig.~\ref{fig:histogram_comparison} c).
\begin{figure}[!htb]
    \centering
    \includegraphics[width=\textwidth]{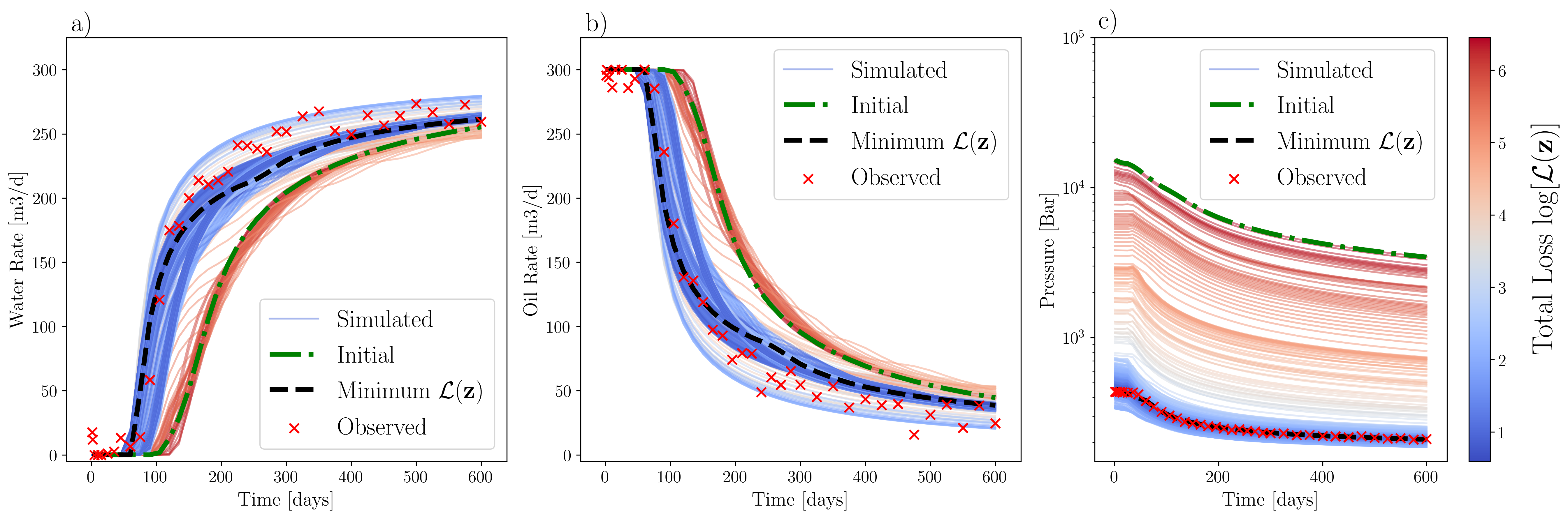}
    \caption{Comparison of the observed dynamic pressure and rate data (red) obtained for a single inversion case. The observed dynamic data from the initial starting model (green) is optimised by honoring flow and well data for 500 iterations. The intermediate pressure and rate data of the 500 iteration steps are colored according to their total loss $\mathcal{L}(\mathbf{z})$.}
    \label{fig:dynamics_flowwells_case_66}
\end{figure}
In figure~\ref{fig:runs_comparison} we have shown that including the flow loss $\mathcal{L}_{flow}(\mathbf{z})$ leads to high variance in the generated distribution of the model properties $\mathbf{m}$ while honoring the observed flow data (Scenario 3, Fig.~\ref{fig:runs_comparison}-c, right), while only constraining samples to honor the well data reduces the variance near the wells, but does not constrain to observed dynamic data (Scenario~2, Fig.~\ref{fig:runs_comparison}-b, right). A smaller injection pressure observed for the samples obtained from scenario 3 compared to those that match only the wells in scenario~2 would indicate a better connected system of channel bodies. After closer inspection of the inverted samples it was found that nearly all models obtained from scenario~3 and 4 are connected (Fig.~\ref{fig:connected_components} and appendix~Fig.~\ref{fig:flow_samples}-\ref{fig:flowwells_samples}). We evaluate for all four scenarios whether any of the clusters of river-channel bodies connects the injection and production wells. Less than half of the models for all scenarios showed a connection between injector and producer wells (Appendix Fig.~\ref{fig:connected_components} - black). After Boolean dilation of the largest cluster and performing the connectivity analysis again, half of the samples that honor the well data only (Scenario 2) show connectivity, whereas all models obtained that were optimised to honor flow (Scenario 3), as well as flow and well data (Scenario~4) form a connected cluster between the injector and producer well (Appendix Fig.~\ref{fig:connected_components} - red). A possible explanation of this result is that due to the finite-difference approximation (two-point flux approximation) used to solve the forward-problem the transmissibilities between two neighboring grid blocks are computed as a weighted average  of the permeability values i.e. a single grid block spacing between two high-permeable channel-bodies does not provide an effective flow barrier.

A single model obtained from optimisation using flow and well losses, while honoring the prior distribution of the latent-variables (Scenario 4), was chosen as an example to highlight the optimisation process for a single realisation. This specific example was chosen as it showed the largest reduction in the total loss $\mathcal{L}(\mathbf{z})$ from the initial starting model to the iteration with the lowest total loss out of 500 optimisation steps. Figure~\ref{fig:functional_flowwells_case_66} shows the evolution of the total loss $\mathcal{L}(\mathbf{z})$ (black), the accuracy in matching the rock-type indicator function $\mathbbm{1}_r$ at the well locations (red), and the prior loss evolution. A highly non-linear optimisation with many local minima can be observed. The accuracy of matching the well data can be observed to quickly reach values above 90\% and staying at 100\% for the entirety of the optimisation process. The smallest total loss for this model is achieved after 214 iterations.

\begin{figure}
    \centering
    \includegraphics[width=\textwidth]{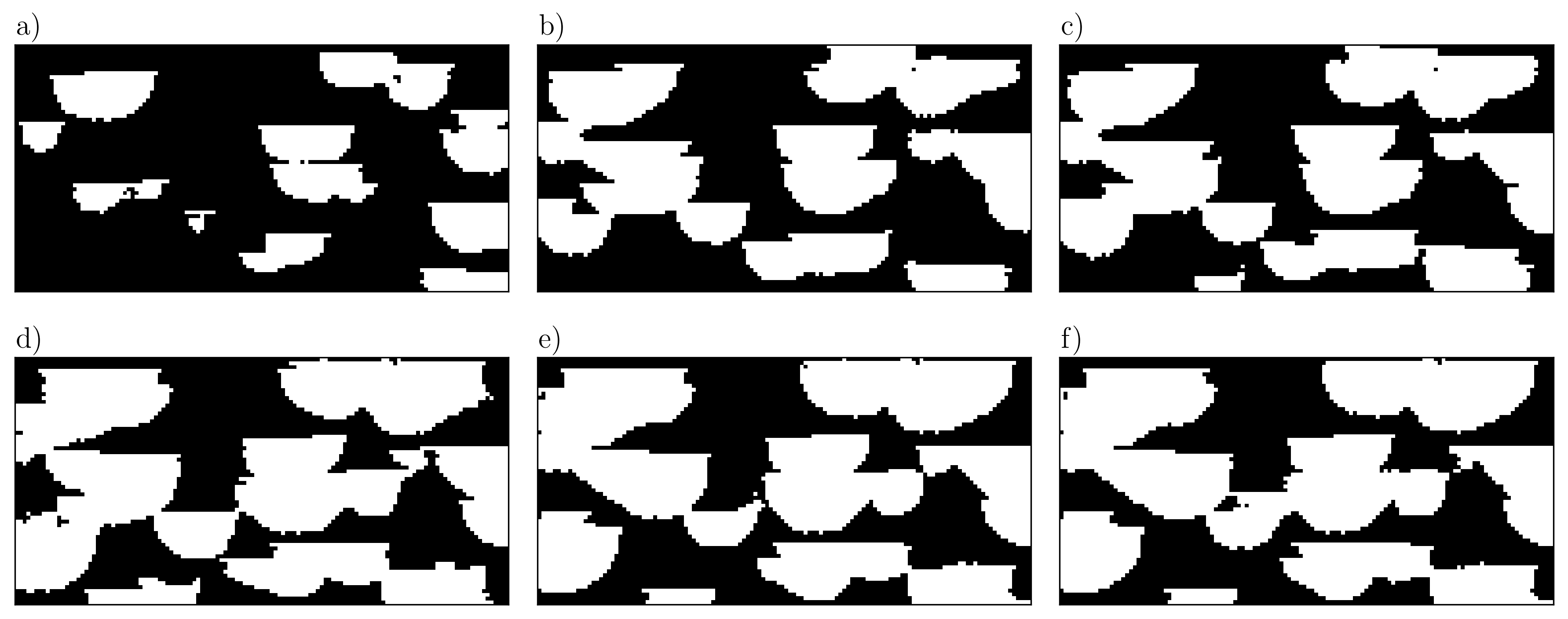}
    \caption{Evolution of the spatial distribution of the rock-type indicator $\mathbbm{1}_r$ for the selected model.}
    \label{fig:facies_flowwells_case_66}
\end{figure}

For the chosen example model, we have recorded the observed dynamic flow behaviour as well as (left column) the distribution of the flow parameters at intermediate steps in the optimisation procedure, where the total loss $\mathcal{L}(\mathbf{z})$ achieved a new local minimum. Figure~\ref{fig:dynamics_flowwells_case_66} shows the evolution of the dynamic flow behaviour in the course of performing the inversion. The model initially shows a very high injection pressure (Fig.~\ref{fig:dynamics_flowwells_case_66} c -blue), which is significantly and continuously reduced to values close to the observed pressure data (red). Similar behaviour is observed for the liquid water and oil rates at the production wells (Fig.~\ref{fig:dynamics_flowwells_case_66} a-b). The solution found by the optimisation strategy (Eq.~\ref{equ:map}) with the lowest error value (Fig.~\ref{fig:dynamics_flowwells_case_66} a-c, black) closely matches the observed dynamic production data and the rock-type indicator function $\mathbbm{1}_r$ at the wells.

The intermediate spatial distributions of the rock-type indicator $\mathbbm{1}_r$ for each of the obtained local minima in the optimisation process are shown in figure~\ref{fig:facies_flowwells_case_66}. Initially the channel-bodies were completely disconnected (Fig.~\ref{fig:facies_flowwells_case_66} a). As the optimisation process progresses more channel-bodies were introduced gradually (Fig.~\ref{fig:facies_flowwells_case_66} b-e) and form a spanning cluster that matches the grid-block data at the wells (Fig.~\ref{fig:facies_flowwells_case_66}~f).
\newpage
\section{Discussion}
\begin{figure}[!htb]
\centering
\begin{subfigure}[b]{0.95\textwidth}
   \includegraphics[width=1\linewidth]{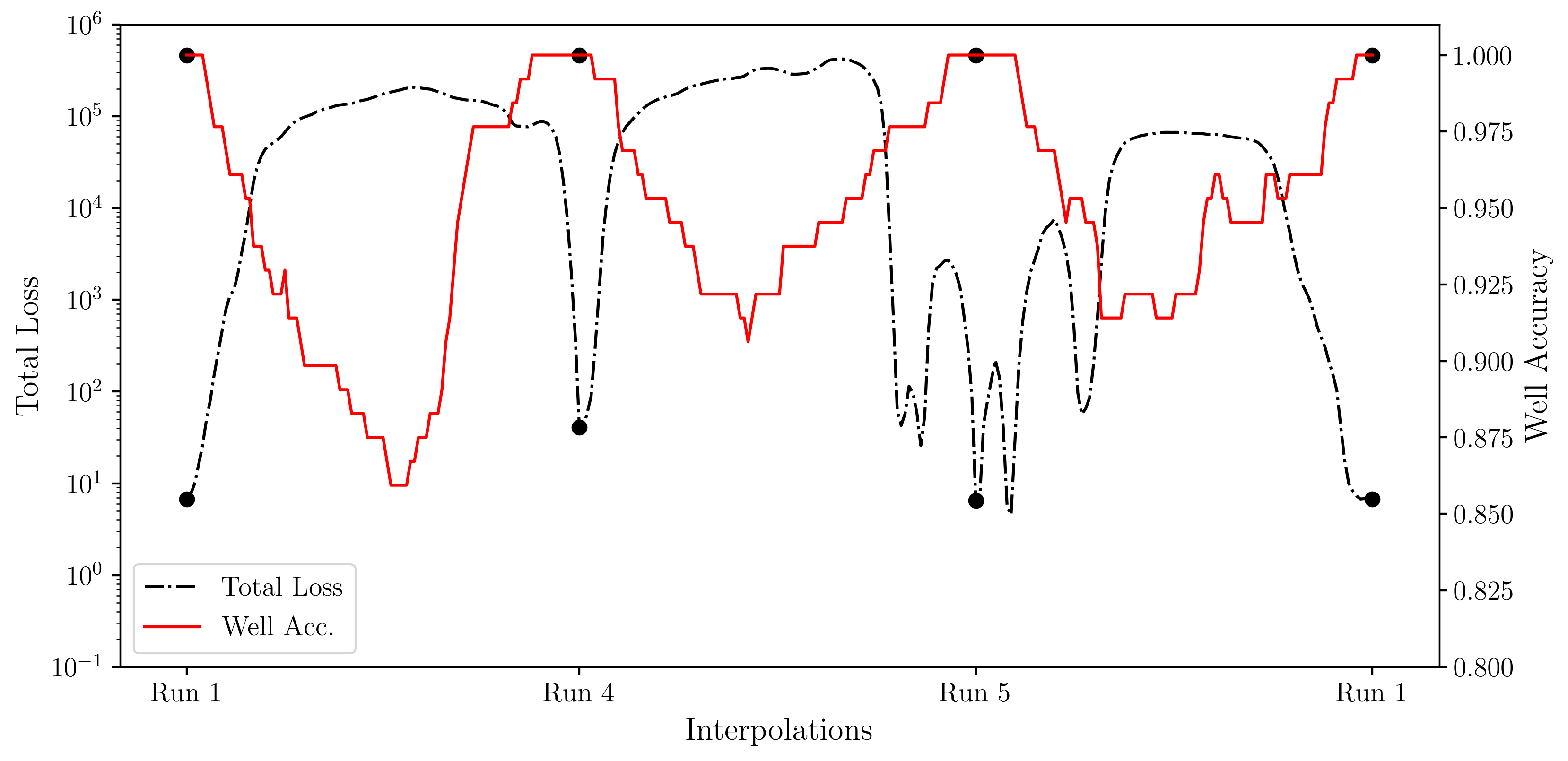}
   \caption{Total Loss $\mathcal{L}(\mathbf{z})$ and well facies accuracy for 100 realisations interpolated between three obtained maxima of the posterior $p(\mathbf{z}|\mathbf{y}_{obs})$.}
   \label{fig:interpolation_total}
\end{subfigure}

\begin{subfigure}[b]{0.95\textwidth}
   \includegraphics[width=1\linewidth]{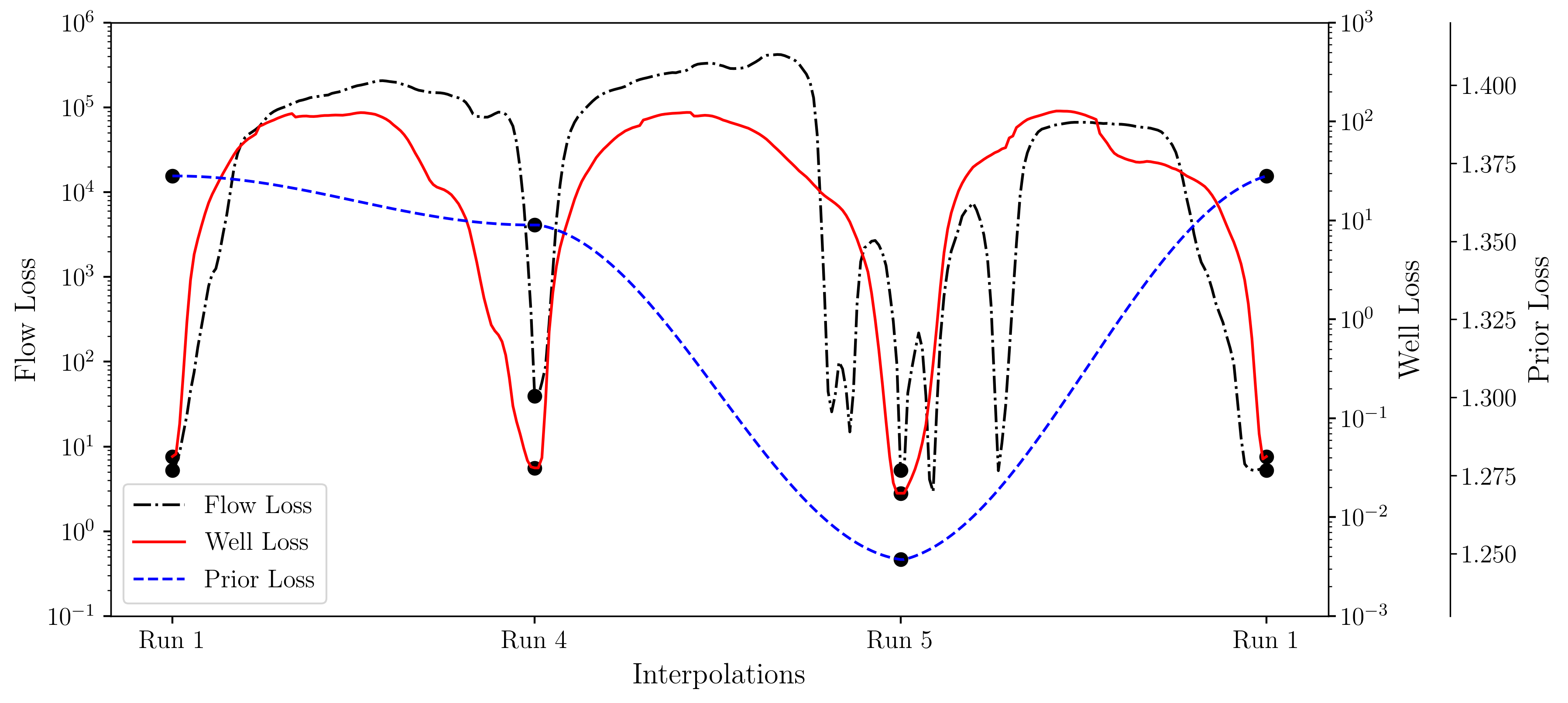}
   \caption{Flow Loss $\mathcal{L}_{flow}(\mathbf{z})$, well loss $\mathcal{L}_{well}(\mathbf{z})$, and prior loss $\mathcal{L}_{prior}(\mathbf{z})$ of the interpolated latent variables between three maxima obtained by optimising the total loss (Scenario 4).}
   \label{fig:interpolation_losses}
\end{subfigure}
\end{figure}

We have shown that solutions of the ill-posed inverse problem of reservoir history matching can be obtained by inversion in the space of a deep generative adversarial network for a synthetic case. The obtained model parameter distributions $\mathbf{m}$ match the dynamic observed flow data $\mathbf{y}_{obs}$ as well as honor the grid-block properties at the wells with adequate solutions found on average in less than 100 optimisation iterations. To evaluate whether the obtained samples of the posterior distribution are local maxima we followed the approach by \citet{zhang2019cyclical} and performed a cyclical spherical linear interpolation \citep{white2016sampling} in the latent space between three obtained realisations. For each interpolated latent vector we have solved the forward model and evaluated the objective functions. Figures~\ref{fig:interpolation_total} and~\ref{fig:interpolation_losses} indicate that the posterior distribution $p(\mathbf{z}|\mathbf{y}_{obs})$ is highly multi-modal, where each of the obtained realisations lie at a minimum of the total loss $\mathcal{L}(\mathbf{z})$ i.e. near a local maximum of the posterior. 
While the obtained samples match the flow and well data it is important to note that in a high-dimensional setting, finding the maxima of the posterior may not be a sufficient description of the full posterior distribution and this needs to be considered when using the obtained history-matched ensemble of reservoir models for forecasts of future reservoir performance. To estimate posterior parameters such as the cumulative production, a harmonic mean averaging approach \citep{green1995reversible,raftery2006estimating} could be used where individual samples are weighted by their corresponding likelihood \citep{zhang2019cyclical}. Furthermore, a number of open challenges are still to be addressed in the context of solving inverse problems with deep generative models such as GANs.

The presented synthetic case is a simplified representation of a system of river-channel bodies. The features encountered in real reservoir systems are more diverse and heterogeneous, and it is possible that many other reservoir features should be included in the training set of generative models. The types of models that may be needed include three-dimensional multi-rock-type models i.e. more than two rock-types in a generated reservoir model. Furthermore, it may be of interest to combine a number of datasets into a single generative model that is conditioned on the type of reservoir architecture $G_{\mathbf{\theta}}(\mathbf{z}, \mathbf{c})$, where $\mathbf{c}$ is a vector of conditional variables that influence the output of the deep generative model using discrete conditionals such as depositional environment e.g. fluvial versus lacustrine environments or using continuous conditioning variables such as reservoir net-to-gross. This type of conditional generative model could be facilitated by making use of architectures such as that proposed by \cite{miyato2018cgans}. Empirically, it has been shown that introducing class-conditional variables to GANs, so-called cGANs, result in improved training and distributional representation behaviour compared to their unconditional counter-parts \citep{gauthier2014conditional}. This may speak for training a single general reservoir model generator compared to creating a library of standalone pre-trained generator networks for each depositional environment. 

An important aspect to be considered here is the evaluation of the generated models in terms of model quality, mode-collapse, and representation of spatial statistics as highlighted in \cite{Mosser17} and \cite{laloy2018training}, where two-point correlation functions are used to quality control the individual generated reservoir models. The use of established GAN quality control measures such as Inception Score (IS) \citep{salimans2016improved} or Frechet Inception Distance (FID) \citep{heusel2017gans} may be challenging due to the induced distribution shift between ImageNet \citep{deng2009imagenet} pre-trained networks and the features observed in typical reservoir architectures. Kernel-based methods such as the maximum mean discrepancy (MMD) have already been successfully applied to train GANs on reservoir models \citep{chan2018parametric} which could be repurposed to quality control generated reservoir models. 

While the synthetic case presented herein has shown that a gradient-based approach to solving the ill-posed inverse problem is feasible, other strategies such as using Markov-Chain Monte-Carlo methods as demonstrated by \cite{LALOY2017387}, may lead to successful inversion results, but at high computational cost. Combinations with other established inversion techniques, such as Ensemble Kalman-Filters \citep{gu2005history, aanonsen2009ensemble} or Ensemble Smoothers (ES-MDA) \citep{emerick2013ensemble} should also be evaluated. The presented approach, together with a set of suitable generative models should be applied to benchmark studies such as the PUNQ or SPE comparative solutions datasets to investigate the applicability of inversion with deep generative models in the presence of measurement errors and noise. While the present study only considers inverting for one model at a time i.e. using a batch-size of one, it is possible to perform inversion using a batch-size greater than one, as in \cite{creswell2018inverting} who show that the gradient contributions to the latent variables can be computed independently for each element of a batch of latent-variables and hence for a set of model parameters $\mathbf{m}$. By computing the numerical solution forward problem and the associated adjoint for each set of model-parameters within a batch, the proposed method could be integrated into existing workflows for ensemble reservoir modeling.  

Using deep latent variable models, such as GANs, to represent the space of solutions for ill-posed inverse problems opens up interesting avenues with regards to the theoretical basis of the ill-posed nature of these problems. Investigation of the existence and stability criteria (Sec.~\ref{sec:introduction}) \citep{tikhonov1977solutions} of deep latent variable inversion schemes could lead to new promising theoretical insights into the convergence behaviour of inversion methods using deep generative models. Furthermore, the geometry of the loss landscape and representation manifold within the space of latent variables may allow for faster convergence and higher quality inversion results. \citet{chen2017metrics} and \citet{arvanitidis2017latent} have investigated the geometric structure of the manifold embedded in the latent-space of deep generative models and \citet{shao2018riemannian} have shown that the manifolds obtained from VAEs are non-linear with near-zero curvature and and hence linear paths in latent space correspond to geodesics on the embedded manifold. This indicates that considering the geometry of the learned manifolds could be accounted for in operations performed in the latent space of deep generative models.

We have shown that GANs may be used as generative models for parameterisation of geological models, and can provide a solution space for ill-posed inverse problems. Nevertheless, GANs  due to their challenging training, evaluation and complex latent-to-parameter-space relationship may not be the optimal reparameterisation choice in the case of ill-posed inverse problems. Deep generative models such as variational \citep{kingma2013auto} and disentangling autoencoders \citep{higgins2017beta,burgess2018understanding}, as well as flow-based generative models, such as RealNVP \citep{dinh2016density} or GLOW \citep{kingma2018glow}, that provide a bijective and invertible mapping may be better suited for the inversion task. This has been demonstrated by \cite{ardizzone2018analyzing} for ill-posed inverse problems.  

\section{Conclusions}
We have presented an application of deep generative models in the context of the ill-posed history matching inverse problem. Based on a two-dimensional synthetic cross-section of a river-channel system we have trained a GAN to represent the prior distribution of the subsurface properties $\mathbf{m}$, permeability $\mathbf{k}$, porosity $\phi$ and the rock-type indicator~$\mathbbm{1}_r$. We find solutions to the two-phase slightly-compressible Darcy flow problem of oil and water systems commonly used to describe the transient flow behaviour in hydrocarbon reservoirs. By incorporating a finite-difference-based numerical simulator with adjoint-state capabilities \citep{krogstad2015mrst} in an end-to-end differentiable framework, we perform inversion using gradient-based optimisation of the mismatch between observed dynamic data (Eq.~\ref{equ:functional}) and grid-block-scale well data (Fig.~\ref{fig:runs_comparison}), while honoring the prior distribution of the latent variables (Eq.~\ref{equ:total_loss_flow_prior}). By using a momentum-accelerated first-order gradient descent scheme \citep{kingma2014adam}, the method converges to a solution despite the highly non-linear and non-convex loss landscape (Fig.~\ref{fig:functional_flowwells_case_66}). Future work will focus on applications to real reservoirs and history matching benchmark studies such as the PUNQ or SPE comparative solutions models, as well as evaluation of other deep latent-variable generative models and their theoretical benefits for finding solutions to ill-posed inverse problems.

\section*{Acknowledgments}
O. Dubrule would like to thank Total S.A. for seconding him as visiting professor at Imperial College London.

\bibliographystyle{agsm}
\bibliography{deepflow}
\newpage
\appendix
\section{Generative Model Architecture and Quality Control}\label{sec:appendix}
\begin{table}[ht]
         \caption{\label{tab:gan_architecture}Generator and discriminator network architectures used to represent the distribution of model parameters $\mathbf{m}$ used in the inversion process (Eq.~\ref{equ:model_parameters}). Binary indicators of geological facies and derived permeability are represented by a bi-variate Gaussian distribution with a hyperbolic tangent activation function used to create the binary output distribution. We use a linear transformation layer to renormalise the output of the GAN from the $(-1, 1)$ interval of the tanh activation to a known range of permeability and porosity values.\\
         Permeability transform parameters: $a=1\times10^{-3} $, $b=1\times10^{-12}$ \\
         Porosity transform parameters: $c=0.3$, $d=0.1$ \\
         Notation for convolutional layers: \\ LayerType(Number of filters),  k=kernel size, s=stride, p=padding. BN=BatchNorm, PS=PixelShuffle}
          \begin{subtable}{.45\textwidth}
              \centering
              {\begin{tabular}[t]{c}
                  \\
                  \toprule
	Latent Variables $z \in \mathbf{R}^{50 \times 2 \times 1}$ \\
                  \midrule
                  Conv2D(512)k3s1p1, BN, ReLU, PSx2 \\
                  \midrule
                  Conv2D(256)k3s1p1, BN, ReLU, PSx2 \\
                  \midrule
                  Conv2D(128)k3s1p1, BN, ReLU, PSx2 \\
                  \midrule
                  Conv2D(64)k3s1p1, BN, ReLU, PSx2 \\
                  \midrule
                  Conv2D(64)k3s1p1, BN, ReLU, PSx2 \\
                  \midrule
                  Conv2D(64)k3s1p1, BN, ReLU, PSx2 \\
                  \midrule
                  Conv2D(2)k3s1p1\\
                  \midrule 
                  $m_0 = Tanh(Channel \ 0)\cdot0.5+0.5$ = $p(\mathbbm{1}_r)$ \\
                  $x_1 = Tanh(Channel \ 1)\cdot0.5+0.5$ \\
                  \midrule
                  $m_1= (a+m_0)\cdot b$ =  permeability $\mathbf{k}$ \\
                  $m_2 = c\cdot x_1+d$ = porosity $\phi$ \\
                  \bottomrule
              \end{tabular}}
              \caption{\label{tab:generator} Multi-Property Generator}
          \end{subtable}
          \hfill
          \begin{subtable}{.45\textwidth}
              \centering
              {\begin{tabular}[t]{c}
              \\
                  \toprule
	Model Parameters $\mathbf{m} \in \mathbf{R}^{2 \times 128 \times 64}$ \\
                  \midrule
                  Conv2D(64)k5s2p2, ReLU \\
                  \midrule
                  Conv2D(64)k5s2p1, ReLU \\
                  \midrule
                  Conv2D(128)k3s2p1, ReLU \\
                  \midrule
                  Conv2D(256)k3s2p1, ReLU \\
                  \midrule
                  Conv2D(512)k3s2p1, ReLU \\
                  \midrule
                  Conv2D(512)k3s2p1, ReLU \\
                  \midrule
                  Conv2D(1)k3s1p1, ReLU \\
                  \bottomrule
              \end{tabular}}
              \caption{\label{tab:discriminator} Discriminator}
          \end{subtable}
          \hspace{2mm}
\end{table}

\begin{table}
\centering
\caption{Fluid and simulation parameters for oil and water phases used in the numerical solution of the two-phase flow forward problem.}\label{tab:simulation_parameters} 
\begin{tabular}{cc|c|c}
              \multicolumn{2}{c|}{Parameter} & Value & Unit \\
              \toprule
\footnotesize Normalised Water Saturation & $\bar{S}_{w}$ & \footnotesize $\frac{S_w-S_{w,cr}}{1-S_{or}-S_{w,cr}}$ &\footnotesize $-$\\
              \midrule
\footnotesize Normalised Oil Saturation & $\bar{S}_{o}$ & \footnotesize $\frac{S_o-S_{o,cr}}{1-S_{wc}-S_{o,cr}}$ &\footnotesize $-$\\
              \midrule
\footnotesize Water, Oil Viscosity & $(\mu_w, \ \mu_o)$ &\footnotesize $(3\times10^{-4}, \ 5\times10^{-3})$ &\footnotesize $Pa\cdot s$ \\
              \midrule
\footnotesize Water, Oil Density & $(\rho_w, \ \rho_o)$ & \footnotesize $(1000, \ 700)$  &\footnotesize $kg\cdot m^{-3}$ \\
              \midrule
\footnotesize Brooks Corey Exponents & $(n_w, \ n_o)$ &\footnotesize $(2, \ 2)$ &\footnotesize $-$ \\
              \midrule
\footnotesize Water, Oil Compressibility & $(c_w, \ c_o)$ &\footnotesize  $(0.0, \ 1\times 10^{-5})$ &\footnotesize $bar^{-1}$ \\
              \midrule
\footnotesize Connate, Init., Crit. Water Saturation & $(S_{wc}, \ S_{w,i}, \ S_{w,cr})$ &\footnotesize $(0.10, \ 0.15, \ 0.15)$ &\footnotesize $-$\\
              \midrule
\footnotesize Residual, Init., Crit. Oil Saturation & $(S_{or}, \ S_{o,i}, \ S_{o,cr})$ &\footnotesize $(0.10, \ 0.15, \ 0.12)$ &\footnotesize $-$ \\
              \midrule
\footnotesize Water Injection Rate & $q_{w,inj}$ & \footnotesize $300$ &\footnotesize $m^3 \cdot day^{-1}$ \\
              \midrule
\footnotesize Bottomhole Pressure & $p_{bhp}$ &\footnotesize $150$ &\footnotesize $bar$ \\
              \midrule
\footnotesize Reference Pressure & $p_{ref}$ &\footnotesize $200$ &\footnotesize $bar$ \\
              \bottomrule
\end{tabular}
\end{table}

\begin{figure}[!htb]
    \centering
    \includegraphics[width=0.5\textwidth]{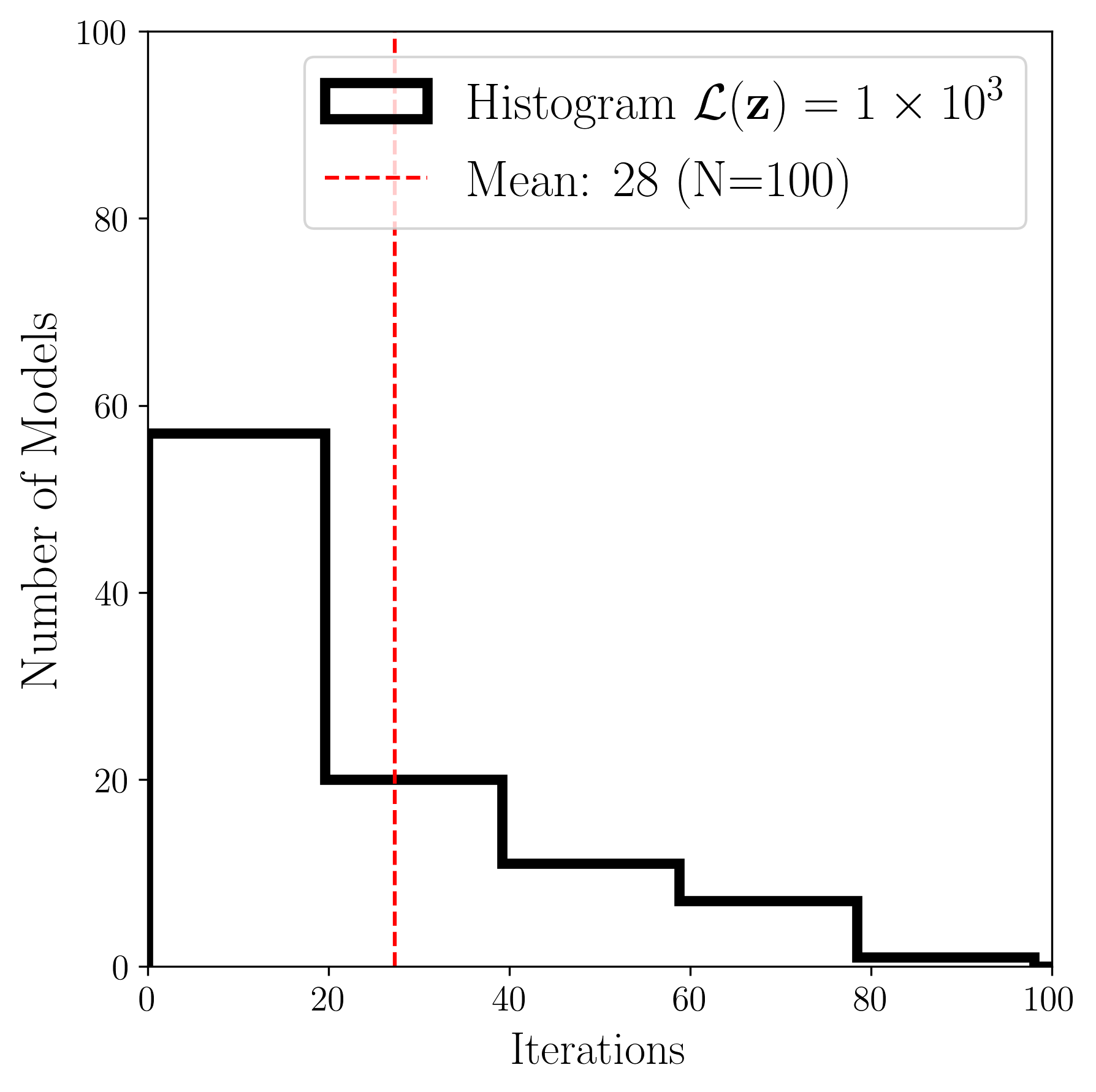}
    \caption{Distribution of the number of optimisation iterations required to reach a total loss $\mathcal{L}(\mathbf{z})$ (Eq.~\ref{equ:functional}) of less than or equal to $1\times10^{3}$ in the case of matching flow and well rock type data (Scenario~4). This threshold corresponds to the lower end of the total-loss distribution of the prior distribution (Fig.~\ref{fig:histogram_comparison}, c). Optimisation converges on average within the first 30 iterations.}
    \label{fig:error_threshold_flowwells}
\end{figure}

\begin{figure}[!htb]
    \centering
    \includegraphics[width=\textwidth]{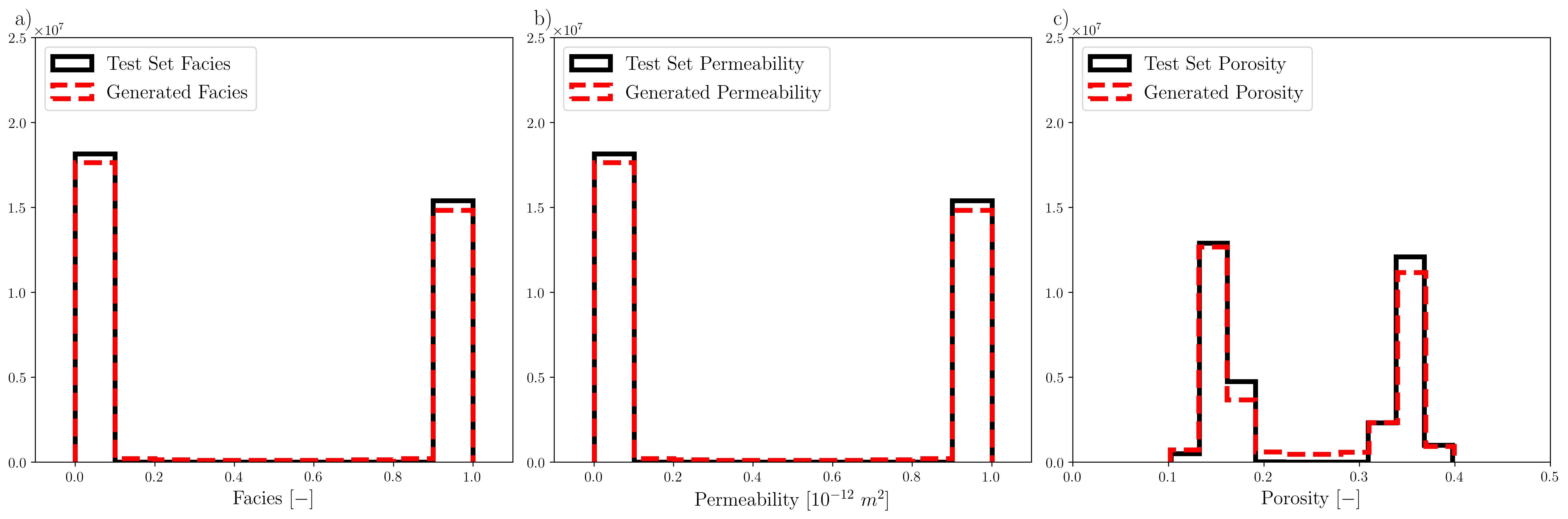}
    \caption{Comparison of the pdfs of $N=1\times10^{4}$ samples of the test set and $N=1\times10^{4}$ samples obtained from the generative network $G_{\mathbf{\theta}}(\mathbf{z})$ trained to represent the river-channel body system. The reservoir rock-type (a) and permeability (b) show a near binary distribution whilst porosity follows a bi-modal pdf. The pdfs of the generated samples match the test set data closely.}
    \label{fig:cdf_properties}
\end{figure}

\clearpage
\section{Inverted Samples}
\begin{figure}[!htb]
    \centering
    \includegraphics[width=0.7\textwidth]{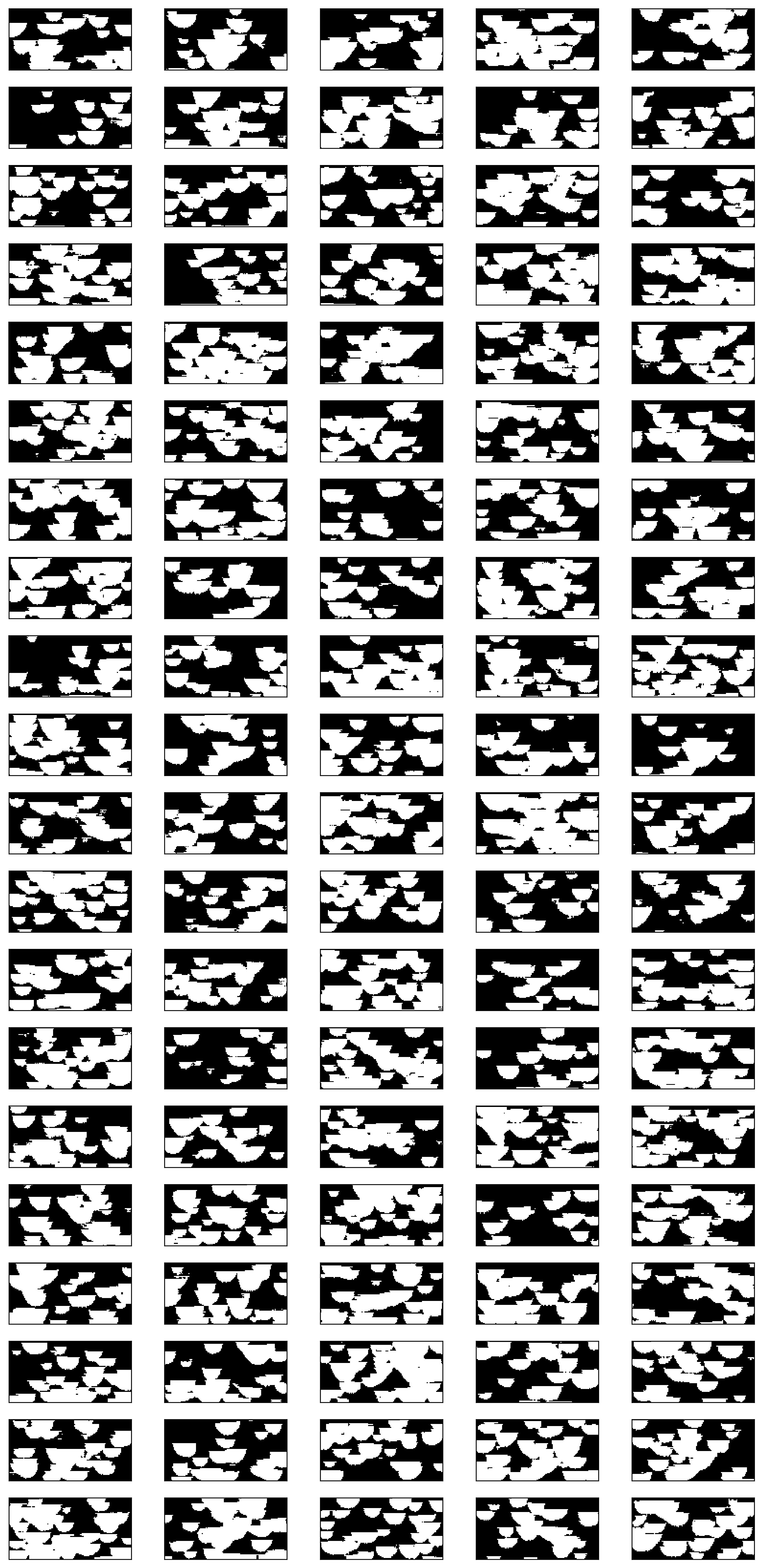}
    \caption{Unconditional samples ($N=100$) obtained from sampling the prior distribution of the generative model (Sec.~\ref{sec:results}, Scenario 1)}
    \label{fig:unconditional_samples}
\end{figure}
\begin{figure}[!htb]
    \centering
    \includegraphics[width=0.7\textwidth]{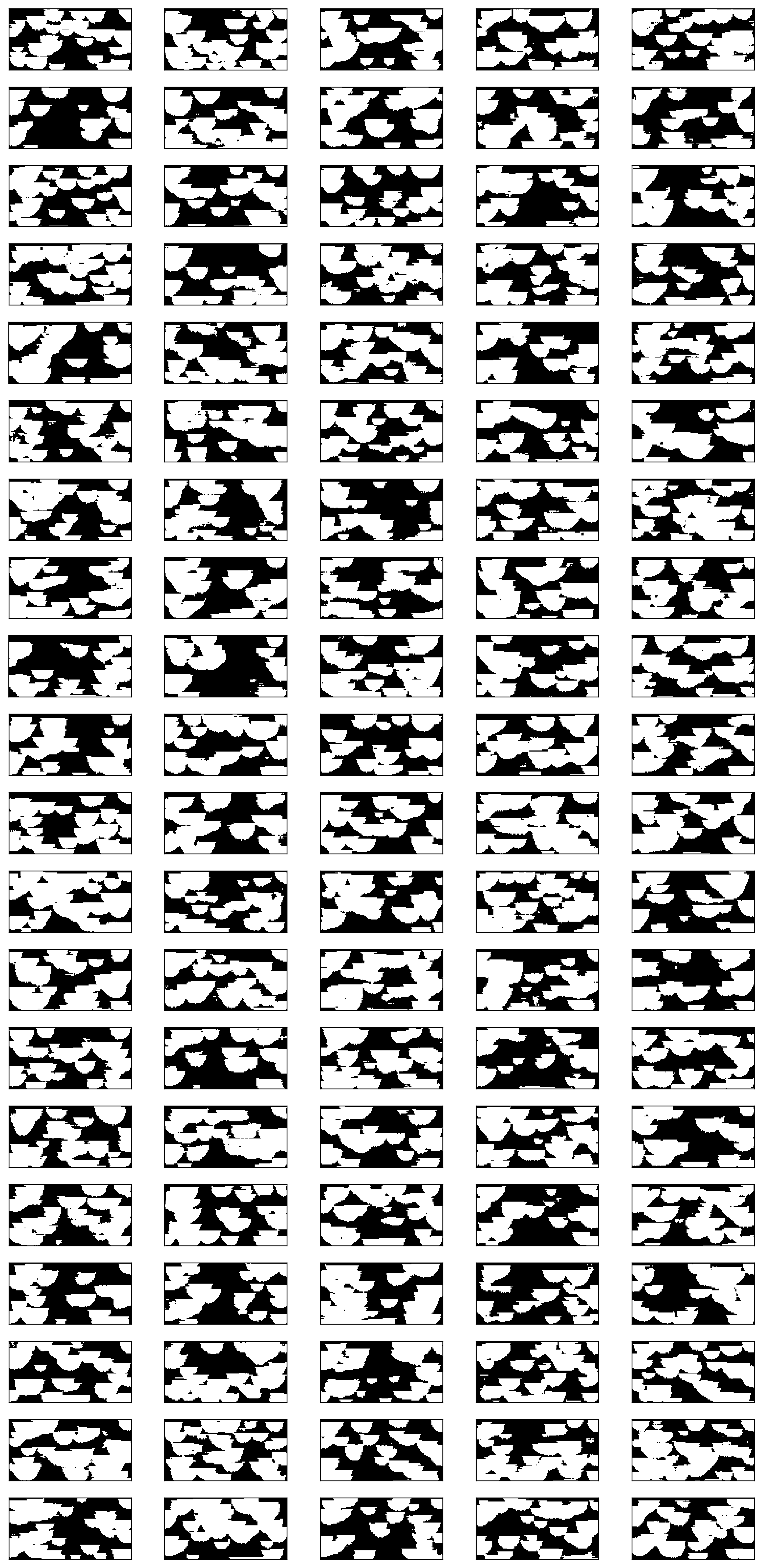}
    \caption{Samples obtained by minimising the sum of the well and prior losses (Sec.~\ref{sec:results}, Scenario 2)}
    \label{fig:well_samples}
\end{figure}
\begin{figure}[!htb]
    \centering
    \includegraphics[width=0.7\textwidth]{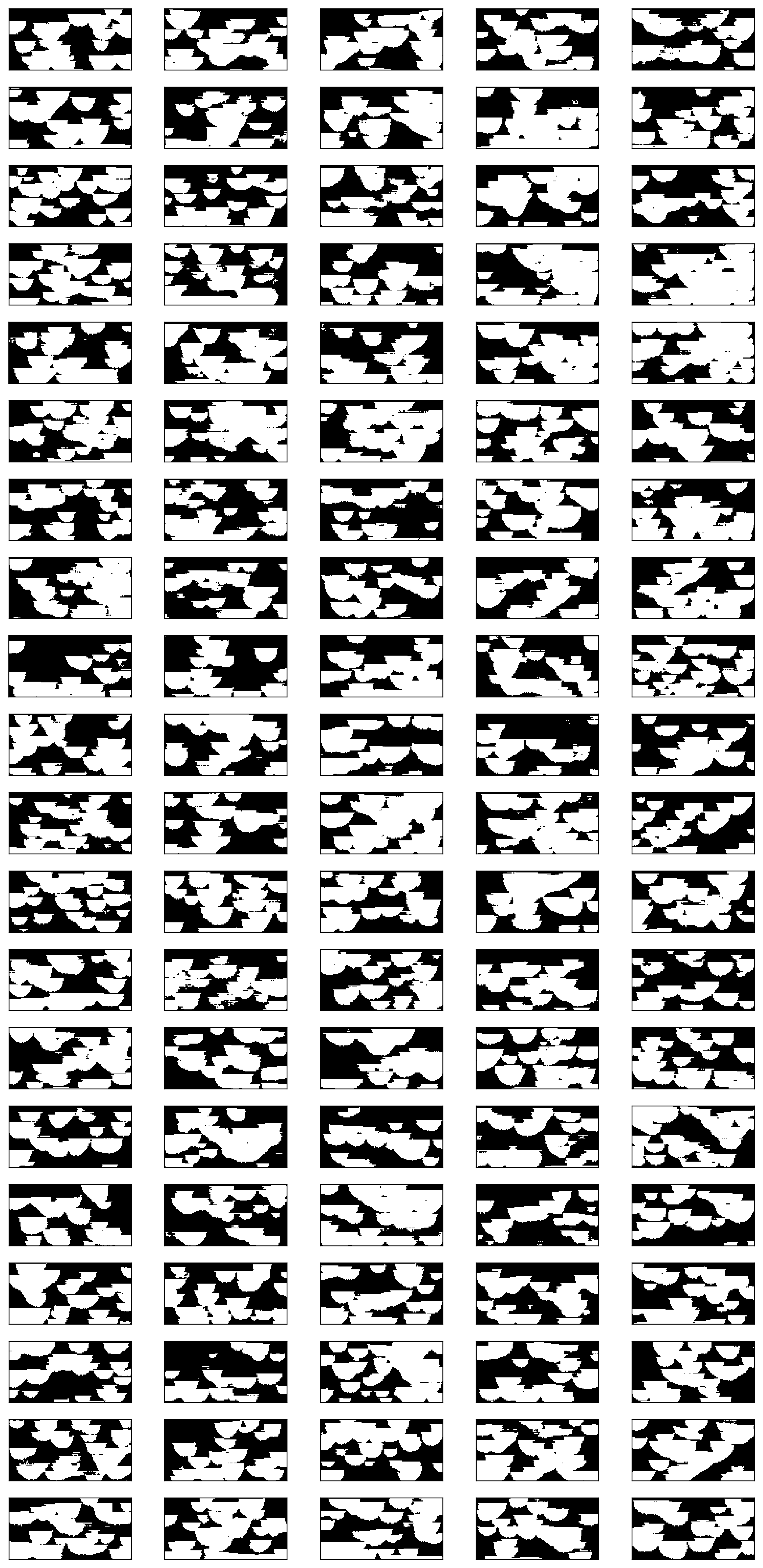}
    \caption{Samples obtained by minimising the sum of the flow and prior losses (Sec.~\ref{sec:results}, Scenario 3)}
    \label{fig:flow_samples}
\end{figure}
\begin{figure}[!htb]
    \centering
    \includegraphics[width=0.7\textwidth]{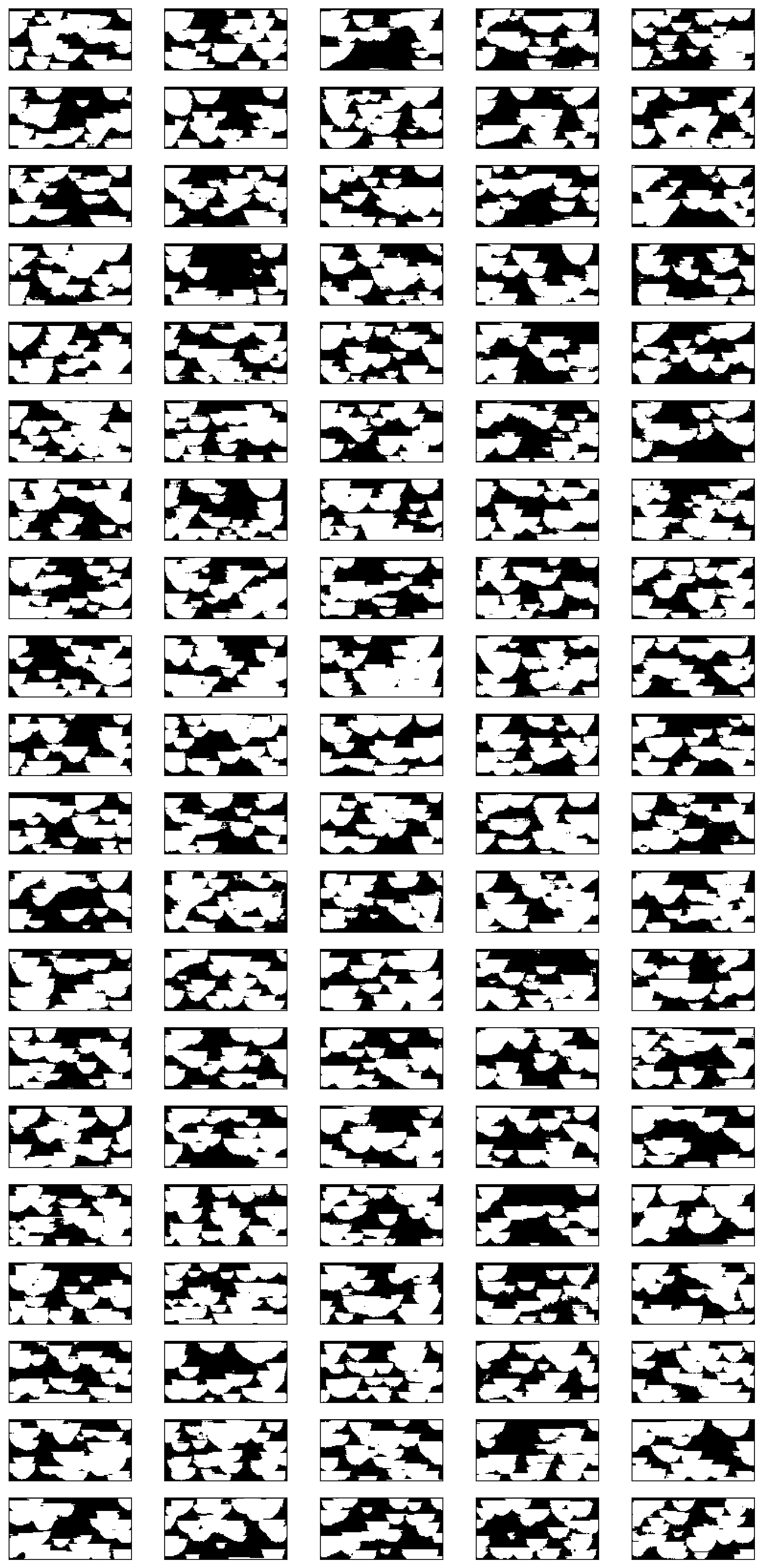}
    \caption{One hundred samples obtained by minimising the sum of the flow, well, and prior losses (Sec.~\ref{sec:results}, Scenario 4)}
    \label{fig:flowwells_samples}
\end{figure}
\end{document}